%% file: main.tex
\documentclass[journal]{IEEEtran}

\usepackage{amsmath, booktabs, etoolbox, graphicx, multirow, url, xcolor}
\usepackage[percent]{overpic}
\usepackage[colorlinks=true, urlcolor=purple, citecolor=black, linkcolor=black]{hyperref}
\usepackage{caption}
\usepackage{anyfontsize}

\makeatletter
\patchcmd{\maketitle}{\newpage}{}{}{}
\makeatother

\begin{document}

\title{Spatially-Aware Evaluation Framework for Aerial LiDAR Point Cloud Semantic Segmentation: Distance-Based Metrics on Challenging Regions}

\let\oldtwocolumn\twocolumn
\renewcommand\twocolumn[1][]{%
  \oldtwocolumn[{#1}{
    \vspace*{-1.5\baselineskip}
    \begin{center}
    \begin{overpic}[width=\textwidth]{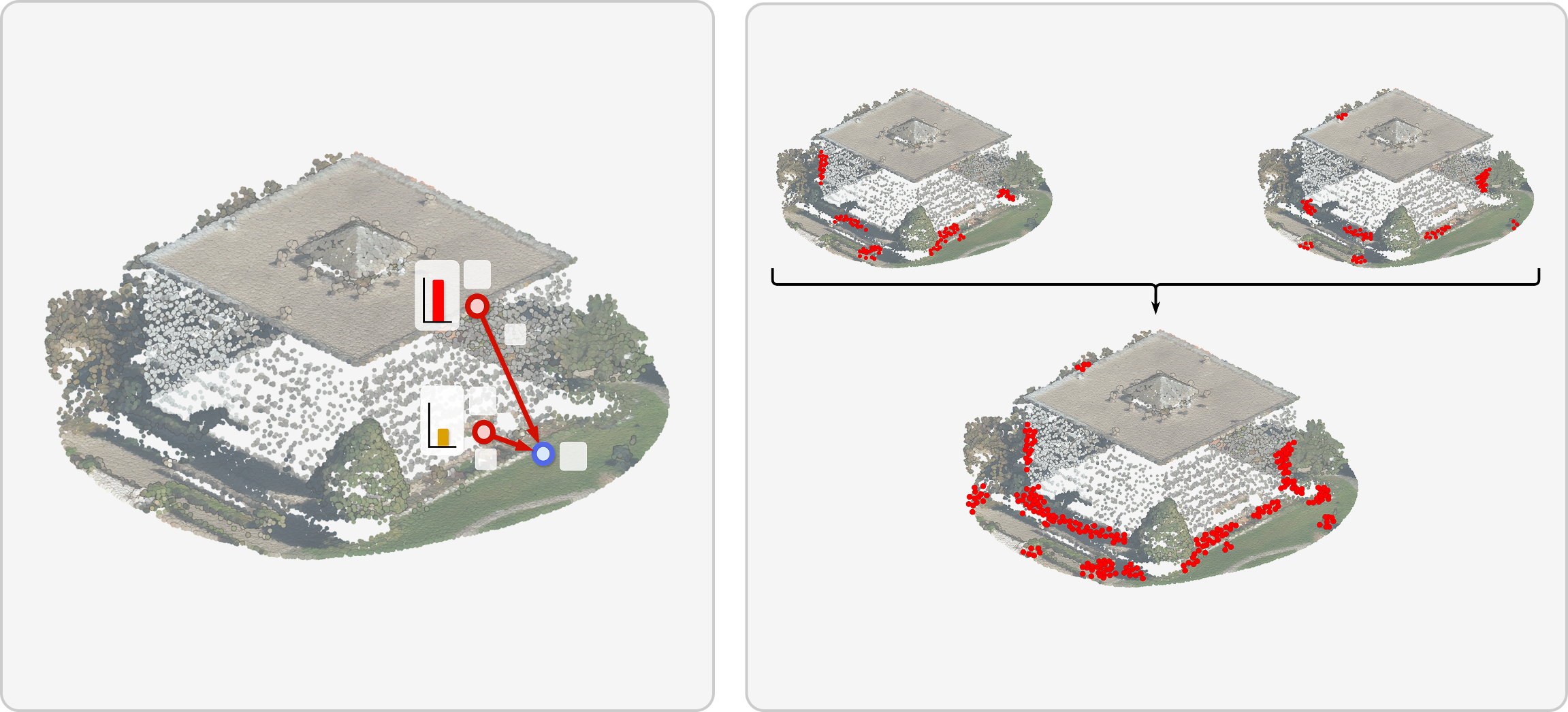} 
        \put(22, 43){\makebox(0,0){\fontsize{9}{11}\selectfont \textbf{(a) Error Distance}}}
        \put(30.4, 27.9){\makebox(0,0){\fontsize{7}{8}\selectfont $A$}}
        \put(30.7, 19.9){\makebox(0,0){\fontsize{7}{8}\selectfont $B$}}
        \put(36.5, 16.3){\makebox(0,0){\fontsize{7}{8}\selectfont $C$}}
        \put(32.9, 24.1){\makebox(0,0){\fontsize{4.5}{4}\selectfont $d_{\!A}$}}
        \put(31, 16.1){\makebox(0,0){\fontsize{4.5}{4}\selectfont $d_{\!B}$}}
        \put(28, 28.2){\makebox(0,0){\fontsize{4}{4}\selectfont $d_A$}}
        \put(28.3, 18.75){\makebox(0,0){\fontsize{4}{4}\selectfont $d_B$}}
        \put(74, 43){\makebox(0,0){\fontsize{9}{11}\selectfont \textbf{(b) Hard Points}}}
        \put(58, 41.5){\makebox(0,0){\fontsize{6}{7}\selectfont $m_1$ misclassifications}}
        \put(73.75, 34.5){\makebox(0,0){\fontsize{12}{14}\selectfont \textbf{\dots}}} 
        \put(89, 41.5){\makebox(0,0){\fontsize{6}{7}\selectfont $m_M$ misclassifications}}
        \put(74, 6.5){\makebox(0,0){\fontsize{7}{9}\selectfont Points misclassified by any model}}
        \put(74, 3.25){\makebox(0,0){\fontsize{8}{10}\selectfont $\displaystyle \mathcal{H} = \bigcup_{i=1}^{M} X_{\text{err}}^{(m_i)}$}}
    \end{overpic}
    \captionof{figure}{
        \textbf{Overview of the proposed Spatially-Aware Evaluation Framework.} 
        \textbf{(a)} Distance-Based error quantification. Two \textit{building} points (A, B) are misclassified as \textit{ground}. Each error is quantified by its distance to the nearest true \textit{ground} point~(C). Therefore, $d_A \gg d_B$ since point A is located on a rooftop (spatially distant from actual ground), while B lies near the building boundary. 
        \textbf{(b)} Hard Points identification. The evaluation focuses on a common subset $\mathcal{H}$ comprising points misclassified by at least one of the compared models, targeting challenging regions such as class boundaries, occlusions, or fine-scale structures where classification is most difficult.
        }
    \label{distance_figure}
    \end{center}
  }]
}

\author{Alex Salvatierra$^{1}$, José Antonio Sanz$^{1}$, Christian Gutiérrez$^{2}$, and Mikel Galar$^{1}$%
\thanks{$^{1}$Department of Statistics, Computer Science and Mathematics and Institute of Smart Cities (ISC), Public University of Navarre (UPNA), Campus de Arrosadía s/n, Pamplona, 31006, Navarre, Spain}%
\thanks{$^{2}$Tracasa Instrumental, Cabárceno, 6, Sarriguren, 31621, Navarre, Spain}%
}

\maketitle

\begin{abstract}
Semantic segmentation metrics for 3D point clouds, such as mean Intersection over Union (mIoU) and Overall Accuracy (OA), present two key limitations in the context of aerial LiDAR data. First, they treat all misclassifications equally regardless of their spatial context, overlooking cases where the geometric severity of errors directly impacts the quality of derived geospatial products such as Digital Terrain Models. Second, they are often dominated by the large proportion of easily classified points, which can mask meaningful differences between models and under-represent performance in challenging regions. To address these limitations, we propose a novel evaluation framework for comparing semantic segmentation models through two complementary approaches. First, we introduce distance-based metrics that account for the spatial deviation between each misclassified point and the nearest ground-truth point of the predicted class, capturing the geometric severity of errors. Second, we propose a focused evaluation on a common subset of \textit{hard points}, defined as the points misclassified by at least one of the evaluated models, thereby reducing the bias introduced by easily classified points and better revealing differences in model performance in challenging regions. We validate our framework by comparing three state-of-the-art deep learning models on three aerial LiDAR datasets. Results demonstrate that the proposed metrics provide complementary information to traditional measures, revealing spatial error patterns that are critical for Earth Observation applications but invisible to conventional evaluation approaches. The proposed framework enables more informed model selection for scenarios where spatial consistency is critical.

The source code is available at {\tt\small \url{https://github.com/arin-upna/spatial-eval}}.
\end{abstract}


\section{Introduction} \label{introduction}

\IEEEPARstart{E}{arth Observation (EO)} has gained significant relevance in recent years, driven by continuous advances in remote sensing technologies, including satellite-based sensors \cite{druschSentinel2ESAsOptical2012}, as well as airborne systems such as aerial LiDAR \cite{wangLiDARPointClouds2018}, capable of capturing high-resolution three-dimensional (3D) information. These developments enable systematic, high-resolution, and temporally frequent data acquisition, supporting applications in environmental monitoring \cite{wulderLidarSamplingLargearea2012}, urban planning \cite{yuAutomatedDerivationUrban2010}, land management \cite{yanUrbanLandCover2015}, and climate analysis \cite{winkerOverviewCALIPSOMission2009}. The growing volume and complexity of EO data make automated processing essential for enabling large-scale geospatial analysis.

Among the various types of data generated in EO, 3D point clouds \cite{liu3DPointCloud2021} have become increasingly relevant due to their capacity to capture detailed surface structure at scale. These data are commonly acquired using Light Detection and Ranging (LiDAR) technology \cite{reutebuchLightDetectionRanging2005}, particularly through Aerial Laser Scanning \cite{zhangEasytoUseAirborneLiDAR2016} systems, which emit laser pulses from airborne platforms to reconstruct the geometry of the terrain and above-ground features. Unlike traditional 2D imagery, LiDAR-derived point clouds provide precise elevation measurements at high spatial density, enabling the extraction of structural information for a wide range of geospatial applications \cite{liuAirborneLiDARGeneration2008}. As a result, LiDAR has become a key tool for generating Digital Terrain Models (DTMs), Digital Surface Models (DSMs), canopy height models, and other 3D products used in forestry, hydrology, infrastructure monitoring, and land cover classification.

Processing and analyzing 3D point clouds derived from aerial LiDAR poses significant challenges due to the irregular, unordered, and sparse nature of the data \cite{kharroubiThreeDimensionalChange2022}. One of the most important tasks in this context is semantic segmentation \cite{zhangReviewDeepLearningBased2019}, which aims to assign a class label to each individual point, enabling a detailed understanding of the scanned environment. Traditional approaches \cite{niemeyerCONDITIONALRANDOMFIELDS2012, zhangSVMBasedClassificationSegmented2013, luSimplifiedMarkovRandom2012} relied on manually designed features and conventional Machine Learning algorithms, which often required extensive preprocessing and were limited in their ability to generalize across different landscapes and acquisition conditions. More recently, Deep Learning (DL) \cite{lecunDeepLearning2015} methods have redefined this task by enabling models to learn relevant patterns directly from raw or minimally processed point cloud data. These models \cite{thomasKPConvFlexibleDeformable2019, huRandLANetEfficientSemantic2020} have achieved strong performance across diverse segmentation scenarios, particularly on large-scale aerial LiDAR datasets \cite{varneyDALESLargescaleAerial2020, yooYUTOSEMANTICLARGE2023}. However, selecting an appropriate model for a specific application may not be straightforward, as standard benchmark metrics may not fully reflect the requirements of the intended use case.

The evaluation of semantic segmentation models on 3D point clouds is typically based on standard classification metrics \cite{zhangReviewDeepLearningBased2019}. The most widely adopted metrics are per-class Intersection over Union (IoU), its macro-average mean Intersection over Union (mIoU), and overall accuracy (OA). IoU measures the overlap between predicted and true class regions, while OA is defined as the proportion of correctly labeled points over the total. These metrics have become standard benchmarks in the literature \cite{guoDeepLearning3D2021a, liDeepLearningLiDAR2021, belloReviewDeepLearning2020a, zhangDeepLearningbased3D2023}, as they provide compact and interpretable indicators of classification performance, and are used extensively for model comparison.

Despite their popularity, these metrics present two important limitations in the context of large-scale aerial LiDAR segmentation. First, they do not account for the spatial severity of misclassifications, treating all errors equally regardless of their geometric deviation from the correct class. As illustrated in Fig.~\ref{distance_figure}a, two \textit{building} points misclassified as \textit{ground} can have vastly different implications: while point B represents a boundary-level error with minimal geometric disruption, point~A constitutes a highly disruptive misclassification located far from any actual ground surface, potentially introducing severe artifacts into derived products such as DTMs. In such cases, models with similar mIoU or OA may differ substantially in their spatial error patterns, with the chosen model potentially committing more disruptive errors that degrade the geometric integrity and practical usability of derived products. Yet, despite the spatial nature of LiDAR data, to the best of our knowledge, no evaluation metric has been proposed to account for the 3D information of each point and quantify the severity of misclassifications.

Second, airborne LiDAR datasets \cite{gaydonFRACTALUltraLargeScaleAerial2024, yooYUTOSEMANTICLARGE2023, varneyDALESLargescaleAerial2020} often contain millions of points per scene, with a large proportion belonging to dominant, easier-to-distinguish classes (e.g., \textit{ground}, \textit{building}) that most models classify correctly regardless of complexity. In this setting, metrics such as OA or mIoU are largely driven by this large proportion of “easy” points, which can lead to similar scores across different models and a reduced ability to distinguish between them in terms of their performance. As a result, these global metrics may mask performance differences in the more ambiguous or structurally complex regions of the scene. Consequently, even when two models achieve similar overall scores, they may exhibit markedly different spatial error patterns that remain hidden to conventional evaluation but can significantly affect the quality of downstream applications. 

To address these limitations, we propose a new spatially-aware evaluation framework for assessing the quality of semantic segmentation models on 3D point clouds. This framework complements traditional metrics by incorporating two fundamental aspects that provide a more comprehensive understanding of model behavior, particularly in contexts where spatial consistency is critical:
\begin{enumerate}
\item A set of \textit{distance-based metrics} that account for the spatial deviation of each error with respect to the ground-truth region of the predicted class. The underlying idea is that errors located far from their corresponding ground truth regions are more severe. By penalizing distant misclassifications more heavily, our metrics encourage predictions that maintain the geometric consistency of the scene, thus improving the quality and usability of the 3D products generated from the segmentation results.
\item An evaluation approach considering only the subset of \textit{hard points}, comprising those misclassified by at least one of the compared models, which focuses the comparison on challenging regions such as class boundaries, occlusions, or structurally complex areas. By restricting the evaluation to this subset, we aim to reduce the bias introduced by large volumes of easily classified points and to better capture meaningful differences among models. Both traditional metrics and our proposed distance-based measures can be evaluated on this subset, providing a more focused and informative comparison of segmentation performance.
\end{enumerate}

To validate the proposed evaluation framework, we conduct an experimental study comparing three state-of-the-art deep learning models for 3D point cloud semantic segmentation: KPConv~\cite{thomasKPConvFlexibleDeformable2019}, RandLA-Net \cite{huRandLANetEfficientSemantic2020}, and Point Transformer V3~\cite{wuPointTransformerV32024}. These models are compared using both traditional and the proposed metrics on three complementary datasets: \textsc{DALES} \cite{varneyDALESLargescaleAerial2020}, \textsc{FRACTAL} \cite{gaydonFRACTALUltraLargeScaleAerial2024}, and a proprietary dataset. This setup allows us to evaluate the consistency and discriminative capacity of the proposed framework across different scenarios and point cloud characteristics.

The remainder of this paper is organized as follows. Section~\ref{related} reviews related work on semantic segmentation of LiDAR point clouds and evaluation metrics. Section~\ref{proposal} introduces our proposed evaluation framework, including distance-based metrics and the definition of hard points. Section~\ref{experimental_framework} describes the experimental setup, detailing the datasets, models, and implementation. Section~\ref{results} presents quantitative results across datasets and models. Section~\ref{discussion} interprets these findings and their implications for evaluation practices. Finally, Section~\ref{conclusions} summarizes the main contributions and outlines future research directions.

\section{Related Work} \label{related}

In this section, we first review recent advances in semantic segmentation of aerial LiDAR point clouds, outlining the specific challenges of this domain (Section~\ref{ssec:review_semantic_segmentation_aerial}). We then examine existing evaluation metrics for 3D point cloud segmentation, highlighting their limitations and motivating the need for spatially-aware evaluation approaches (Section~\ref{ssec:review_metrics}).

\subsection{Semantic Segmentation of Aerial LiDAR Point Clouds} \label{ssec:review_semantic_segmentation_aerial}

Semantic segmentation of 3D point clouds has become a key task in 3D scene understanding, with a wide range of applications in robotics, autonomous driving, and geospatial analysis. Since the introduction of PointNet \cite{charlesPointNetDeepLearning2017}, which pioneered the direct processing of unordered point sets, numerous deep learning architectures \cite{guoDeepLearning3D2021a, liDeepLearningLiDAR2021} have been proposed to improve segmentation performance and computational efficiency. These include point-wise networks such as PointNet++ \cite{qiPointNetDeepHierarchical2017} and RandLA-Net \cite{huRandLANetEfficientSemantic2020}, convolution-based methods like KPConv \cite{thomasKPConvFlexibleDeformable2019} and PointCNN \cite{liPointCNNConvolutionXTransformed2018}, and transformer-based approaches including Superpoint Transformer \cite{robertEfficient3DSemantic2023} and the Point Transformer family \cite{zhaoPointTransformer2021, wuPointTransformerV22022, wuPointTransformerV32024}.

The most common datasets for evaluating 3D semantic segmentation involve indoor and ground-based scenarios, such as S3DIS \cite{armeni3DSemanticParsing2016}, ScanNet \cite{daiScanNetRichlyAnnotated3D2017}, Semantic3D \cite{hackelSEMANTIC3DNETNEWLARGESCALE2017}, and SemanticKITTI \cite{behleySemanticKITTIDatasetSemantic2019}, which vary significantly in acquisition technology, from RGB-D sensors to mobile and terrestrial LiDAR, and in spatial structure and scene complexity. 

More recently, large-scale aerial LiDAR datasets that enable the training and evaluation of aerial LiDAR semantic segmentation models have been made available, including \textsc{DALES} \cite{varneyDALESLargescaleAerial2020}, \textsc{FRACTAL} \cite{gaydonFRACTALUltraLargeScaleAerial2024}, YUTO \cite{yooYUTOSEMANTICLARGE2023} and ECLAIR \cite{melekhovECLAIRHighFidelityAerial2024}. These point clouds present unique challenges including irregular point density, massive spatial scales with millions of points per scene, significant elevation variations, and multi-scale objects ranging from fine power lines to large buildings. It is also common to find severe class imbalance, with dominant ground and vegetation classes vastly outnumbering minority categories such as vehicles or poles. Additionally, the nadir acquisition angle results in reduced point density on vertical structures compared to horizontal surfaces. Fortunately, although many architectures were not originally designed for airborne data, several have shown strong performance on these benchmarks, highlighting their generalization capacity from indoor and ground-based scenarios \cite{guoDeepLearning3D2021a}.

\subsection{Evaluation Metrics for 3D Point Cloud Segmentation} \label{ssec:review_metrics}

The evaluation of 3D segmentation models is typically based on standard classification metrics such as overall accuracy (OA), mean Intersection over Union (mIoU), and per-class IoU. These metrics offer a compact and interpretable summary of model performance and are widely used in comparative studies \cite{guoDeepLearning3D2021a, liDeepLearningLiDAR2021, singhDeepLearningbasedSemantic2024, zhangDeepLearningbased3D2023}. 

Given a point cloud with $N$ points and a set of $N_c$ semantic classes $\mathcal{C} = \{1, \ldots, N_c\}$, let $y_i \in \mathcal{C}$ and $\hat{y}_i \in \mathcal{C}$ denote the ground-truth and predicted labels for point $p_i$ ($i = 1, \ldots, N$), respectively. For each class $c \in \mathcal{C}$, these metrics are defined as follows:

\begin{equation}
\label{eq:oa}
\mathrm{OA} = \frac{\sum_{c=1}^{N_c} TP_c}{N}
\end{equation}

\begin{equation}
\label{eq:iou}
\mathrm{IoU}_c = \frac{TP_c}{TP_c + FP_c + FN_c}
\end{equation}

\begin{equation}
\label{eq:miou}
\mathrm{mIoU} = \frac{1}{N_c} \sum_{c=1}^{N_c} \mathrm{IoU}_c
\end{equation}

where $TP_c$, $FP_c$, and $FN_c$ denote true positives, false positives, and false negatives for class $c$, respectively.

While these metrics provide a global picture of classification performance, they do not account for the spatial nature of point cloud data, limiting their ability to provide a finer-grained characterization of segmentation errors. This limitation is not unique to point cloud segmentation, and distance-based evaluation has been explored in related domains to incorporate spatial information into error quantification: 3D scene reconstruction uses geometric accuracy \cite{knapitschTanksTemplesBenchmarking2017}, image segmentation employs Hausdorff distance \cite{huttenlocherComparingImagesUsing1993} and surface distances \cite{GoogledeepmindSurfacedistance2018} (though not widely adopted), and object detection incorporates spatial relationships through metrics like Localization Recall Precision \cite{oksuzLocalizationRecallPrecision2018} and Distance-IoU \cite{zhengDistanceIoULossFaster2020}. 

Beyond distance-based approaches, related work has also explored assigning different costs or weights to misclassifications, though typically based solely on classification rather than the spatial information of individual errors. For instance, cost-sensitive learning \cite{elkanFoundationsCostsensitiveLearning2001} and weighted metrics \cite{paredesLearningWeightedMetrics2006} aim to assign error importance, but they fail to capture individual error severity and defining appropriate weights remains a non-trivial task. Other approaches include adaptive losses such as Focal Loss \cite{linFocalLossDense2018}, which emphasize hard misclassifications while down-weighting easy ones, and hierarchical evaluation frameworks \cite{sunHierarchicalTextClassification2001,dekelLargeMarginHierarchical2004}, where the penalty of an error grows with the semantic distance between the true and predicted classes. Therefore, none of these existing approaches adequately address the specific challenges of 3D point cloud segmentation.

In this work, we aim to address these limitations by introducing a spatially-aware evaluation framework that incorporates geometric error severity and focuses on challenging regions. Our framework complements existing metrics by providing more informative model comparisons, especially in the context of EO where spatial coherence is essential for the generation of reliable geospatial products.

\section{Proposal: Spatially-Aware Evaluation Framework} \label{proposal}

As discussed earlier, existing metrics for 3D point cloud segmentation fail to capture the spatial magnitude of misclassifications and are often dominated by large volumes of easily classified points. To overcome these limitations, we propose an evaluation framework that integrates two complementary perspectives: distance-based metrics that quantify the geometric deviation of misclassified points, emphasizing the impact of spatially disruptive errors; and a hard points-based evaluation that restricts comparison to a shared subset of challenging points, thereby reducing the influence of trivial classifications.

\subsection{Distance-Based Metrics}

The proposed distance-based metrics consider how far each classified point lies from the nearest ground-truth instance of its predicted class, providing a spatial quantification of error severity. 

As illustrated in Fig.~\ref{distance_figure}a, this approach allows distinguishing between errors with minimal geometric deviation (e.g., boundary-level misclassifications) and those representing severe spatial disruptions. Unlike conventional metrics such as IoU or OA, which focus solely on classification accuracy, these metrics explicitly capture the geometric severity of misclassifications. With this idea in mind, distance-based metrics are defined as follows.

Let $X = \{ p_i \}_{i=1}^N$ be the set of all points, where each point $p_i$ is associated with a ground-truth label $y_i$ and a predicted label $\hat{y}_i^{(m)}$ from model $m$. For any class $c$, denote
\begin{equation} \label{eq:def_Xc}
X_{c}=\{\,p_i\in X \mid y_i=c\,\},
\end{equation}

\begin{equation} \label{eq:def_Xyhat}
X_{\hat{y}=c}=\{\,p_i\in X \mid \hat y_i^{(m)}=c\,\}.
\end{equation}

For each point $p_i \in X$, we compute its distance to the nearest ground-truth point whose ground truth matches the predicted class:
\begin{equation} \label{eq:dist_raw}
d_i^{\text{raw}} = \min_{p_j \in X_{y=\hat{y}_i}} \| p_i - p_j \| .
\end{equation}

where $\|\cdot\|$ denotes the Euclidean distance in 3D space. Note that for correctly classified points, $d_i^{\text{raw}} = 0$ since the point itself belongs to the set of ground-truth points with class $\hat{y}_i$. This distance measures the severity of the misclassification. 

However, extremely large distances may appear, providing limited additional insight and disproportionately influencing average values. For instance, a single \textit{vegetation} point incorrectly labeled as \textit{building} in a forested area may lie hundreds of meters from the nearest true \textit{building}, thereby exaggerating the perceived severity of an isolated error. To prevent these types of outliers from dominating the metrics, we establish a class-specific threshold $\tau_c$ (see Table~\ref{tab:thresholds} in Section~\ref{experimental_framework}) to limit their influence:
\begin{equation} \label{eq:dist_clipped}
d_i = \min\big(d_i^{\text{raw}},\, \tau_{\hat{y}_i}).
\end{equation}

Using these clipped distances computed for each point, we can now define a per-class metric summarizing the severity of points classified as class $c$. The Mean Distance Error (MDE) for class $c$ is the average of the clipped distances for points predicted as class $c$:
\begin{equation} \label{eq:mde_c}
\mathrm{MDE}_c = \frac{1}{|X_{\hat{y}=c}|} 
\sum_{p_i \in X_{\hat{y} = c}} d_i.
\end{equation}

These per-class metrics can be macro-averaged to obtain a single global indicator over all $N_c$ classes:
\begin{equation} \label{eq:mmde}
\mathrm{mMDE} = \frac{1}{N_c} \sum_{c=1}^{N_c} \mathrm{MDE}_c .
\end{equation}

Additionally, these distances can be used for computing additional metrics providing insights about the distribution of spatial deviations. Let us first define the sets of distant points (Eq.~\ref{eq:distant}), misclassified points (Eq.~\ref{eq:err_set}) and non-distant errors (Eq.~\ref{eq:near_set}) for class $c$:

\begin{equation} \label{eq:distant}
X_{\hat{y}=c}^{\text{distant}} = \{\, p_i \in X_{\hat{y}=c} \mid d_i^{\text{raw}} > \tau_{\hat{y}_i} \,\},
\end{equation}

\begin{equation} \label{eq:err_set}
X_{\hat{y}=c}^{\text{err}} = \{\, p_i \in X_{\hat{y}=c} \mid \hat{y}_i^{(m)} \neq y_i \,\},
\end{equation}

\begin{equation} \label{eq:near_set}
X_{\hat{y}=c}^{\text{near}} = X_{\hat{y}=c}^{\text{err}} \setminus X_{\hat{y}=c}^{\text{distant}}.
\end{equation}

Using the previous definitions we can compute the proportion of distant errors among all misclassified points (Eq.~\ref{eq:rho_c}), which quantifies how frequently a model produces distant misclassifications for class $c$, and  the mean distance among non-distant errors (Eq.~\ref{eq:mu_c}) for class $c$, which provides additional information about how fine-grained the classification is near the true class boundaries among non-distant misclassifications:
\begin{equation} \label{eq:rho_c}
\rho_c = \frac{|X_{\hat{y}=c}^{\text{distant}}|}{|X_{\hat{y}=c}^{\text{err}}|},
\end{equation}

\begin{equation} \label{eq:mu_c}
\mu_c = \frac{\sum_{p_i \in X_{\hat{y}=c}^{\text{near}}} d_i}{|X_{\hat{y}=c}^{\text{near}}|}.
\end{equation}

When applied to aerial LiDAR datasets containing millions of points, the large proportion of correctly classified points, each contributing zero distance, may dilute the average mMDE and obscure meaningful spatial error patterns. To address this limitation, we propose evaluating distance-based metrics on a common subset of challenging points, as described in the next section.

A possible alternative would be to restrict the evaluation to misclassified points only. However, because each model yields a different set of misclassified points, this approach would not allow for a fair direct comparison between models. Still, distance metrics computed exclusively on misclassified points may provide useful complementary information when used alongside traditional classification metrics such as IoU.

\subsection{Hard Points-Based Model Evaluation}

In addition to capturing spatial error severity, our framework also mitigates the effect of the large number of trivially classified points in airborne LiDAR scenes. These points, often belonging to dominant classes such as \textit{ground} or \textit{building}, can inflate global metrics and obscure performance differences in more ambiguous or structurally complex regions. To overcome this limitation and obtain more interpretable distance-based measures, we restrict evaluation to a common subset of hard points, $\mathcal{H} \subseteq X$, where at least one of the evaluated models fails. 

\begin{equation} \label{eq:hard_points}
\mathcal{H} = \left\{ p_i \in X \,\middle|\, \exists m \in \mathcal{M}, \hat{y}_i^{(m)} \neq y_i \right\}
\end{equation}

where $\mathcal{M}$ is the set of models being compared.

As shown in Fig.~\ref{distance_figure}b, this subset comprises challenging regions such as class boundaries, occlusions, or structurally complex areas where models struggle to produce consistent predictions and where robustness and discriminative ability can be assessed more clearly.

In practice, all metrics---both traditional (IoU, OA) and distance-based (MDE, $\rho$, $\mu$)---are computed using the same definitions as on the full test set, but restricted to the subset of hard points $\mathcal{H}$. Importantly, distance computations (Eq.~\ref{eq:dist_raw}) still reference all ground-truth points in $X$, rather than only those in $\mathcal{H}$, so that spatial severity is evaluated against the complete scene geometry. In the experiments, we report results on both the full test set and $\mathcal{H}$, enabling a direct comparison of the complementary insights offered by each perspective.

\section{Experimental Framework} \label{experimental_framework}

This section presents the experimental setup used to validate the proposed framework. We first describe the three aerial LiDAR datasets considered in this work (Section~\ref{ssec:datasets}), then the deep learning models selected for evaluation (Section~\ref{ssec:models}), and finally the training configuration and implementation details (Section~\ref{ssec:training_details}).

\subsection{Datasets} \label{ssec:datasets}

We conduct our evaluation on three large-scale aerial LiDAR datasets: two well-known public benchmarks \textsc{DALES} \cite{varneyDALESLargescaleAerial2020} and \textsc{FRACTAL} \cite{gaydonFRACTALUltraLargeScaleAerial2024}, and a proprietary dataset with complementary characteristics. These datasets cover a variety of geographic contexts, labeling schemes and dataset sizes, enabling a comprehensive assessment of the proposed evaluation framework under diverse spatial conditions and class distributions.

\vspace{6pt}\noindent\textbf{\textsc{DALES}} \cite{varneyDALESLargescaleAerial2020} is a large-scale aerial LiDAR dataset covering 10 km² of terrain in the city of Surrey in British Columbia, Canada. The dataset contains 505 million labeled points distributed across eight semantic classes: ground, vegetation, cars, trucks, power lines, fences, poles, and buildings. The data covers four distinct scene types: commercial, urban, rural, and suburban areas. Data was collected using a Riegl Q1560 dual-channel system at 1300 meters altitude with 400\% minimum overlap, achieving an average density of 50 points per square meter.

\vspace{6pt}\noindent\textbf{\textsc{FRACTAL}} \cite{gaydonFRACTALUltraLargeScaleAerial2024} is an ultra-large-scale aerial LiDAR dataset spanning 250 km² across five distinct regions of southern France. The dataset contains over 9.2 billion labeled points distributed into seven semantic classes: ground, vegetation, building, water, bridge, permanent structure, and other. The original data was acquired using various sensors (Leica, Riegl, and Teledyne/Optech) with a target density of 10 pulses per square meter, achieving an average of 40 points per square meter. Data was sampled from the French Lidar HD program using a targeted sampling strategy that explicitly concentrates rare classes and challenging landscapes across diverse environments including urban, rural, mountainous, and coastal areas.

\vspace{6pt}\noindent\textbf{\textsc{Tracasa-PNA20}} is a large-scale proprietary aerial LiDAR dataset covering 4 km² in the Pamplona metropolitan area in Navarra, Spain. The dataset contains 140 million labeled points distributed across five semantic classes: ground, low vegetation, medium/high vegetation, building, and vehicle. Data was collected using a Leica CityMapper-2 hybrid airborne sensor at 1000 meters altitude, achieving an average density of 50 points per square meter. The data encompasses diverse landscape types including urban, industrial, rural, and semi-rural areas.

\vspace{2mm}

\subsection{Models} \label{ssec:models}

We evaluate three state-of-the-art deep learning models for 3D semantic segmentation of aerial LiDAR point clouds: KPConv, RandLA-Net, and Point Transformer V3. These models were selected for their widespread use, architectural diversity and strong performance in large-scale outdoor segmentation tasks.

\vspace{6pt}\noindent\textbf{KPConv} \cite{thomasKPConvFlexibleDeformable2019} is a convolution-based method that employs kernel point convolutions to adapt the traditional 2D convolution operator to irregular 3D point clouds. The method uses a set of kernel points in 3D space, where each kernel point has associated weight matrices that are applied to input points based on their spatial proximity. The convolution operation aggregates features from neighboring points using linear interpolation weights that decrease with distance.

\vspace{6pt}\noindent\textbf{RandLA-Net} \cite{huRandLANetEfficientSemantic2020} is a point-wise method designed for efficient processing of large-scale point clouds. It relies on random sampling to reduce computational cost, enabling fast processing of scenes with millions of points. To preserve geometric detail despite random downsampling, it introduces the Local Feature Aggregation module, which enriches point features through spatial encoding and attention-based aggregation within local neighborhoods.

\vspace{6pt}\noindent\textbf{Point Transformer V3 (PTv3)} \cite{wuPointTransformerV32024} abandons the strict permutation-invariant treatment of point sets by serializing point clouds into ordered sequences and applying patch-wise attention. This replaces costly KNN-based neighborhoods and heavy positional encodings with a simpler, more efficient scheme that greatly expands the receptive field.

\subsection{Implementation Details} \label{ssec:training_details}

To ensure a rigorous and reproducible evaluation, all experiments were conducted under a unified pipeline with consistent preprocessing, training configurations, and evaluation protocols across all datasets and models.

\paragraph*{Data Processing}

All datasets are organized into $50 \times 50$~m tiles. \textsc{DALES} and \textsc{Tracasa-PNA20} scenes were tiled into $50 \times 50$~m patches with $25$~m overlap, while \textsc{FRACTAL} is pre-tiled at $50 \times 50$~m. \textsc{FRACTAL} uses official splits comprising 80,000 training tiles, 10,000 validation tiles, and 10,000 test tiles. For \textsc{Tracasa-PNA20} and \textsc{DALES}, 10\% of the training tiles were reserved for validation, with the remaining 90\% used for training. Original test sets were preserved in all cases, and test labels were never used for model selection.

\textbf{\textsc{FRACTAL}} employs the official transformations from the authors' repository \cite{gaydonFRACTALUltraLargeScaleAerial2024}, which include voxel-based grid sampling (0.25~m) and fixed-scale coordinate normalization. \textbf{\textsc{Tracasa-PNA20}} and \textbf{\textsc{DALES}} use tile-relative normalization with coordinates centered and scaled to $[-0.5, 0.5]$. Regarding input features, \textsc{FRACTAL} and \textsc{Tracasa-PNA20} provide $(x,y,z)$, intensity, RGB, return metadata, NIR, and NDVI. \textsc{DALES} provides only $(x,y,z)$ and intensity (4 channels).
For the evaluation of the test set, the same preprocessing is applied. For \textsc{Tracasa-PNA20} and \textsc{DALES}, predictions from overlapping tiles are merged by averaging per-class probabilities at inference.

\paragraph*{Training Protocol}

All experiments were conducted on an NVIDIA RTX 6000 Ada GPU with 48~GB of VRAM. Models were implemented using publicly available source code\footnote{KPConv: {\tt\small \url{https://github.com/torch-points3d/torch-points3d}}; RandLA-Net: {\tt\small \url{https://github.com/IGNF/myria3d}}; PTv3: {\tt\small \url{https://github.com/pointcept/pointcept}}.} and trained under the same experimental conditions to ensure comparability. Their configurations followed recommended hyperparameter settings from the original publications, with minor adjustments to accommodate the spatial scale and density of aerial LiDAR data.

Batch sizes were adjusted per dataset and architecture (Table~\ref{tab:batch_sizes}) to account for their varying memory requirements and all models use cross-entropy loss. Training proceeded until validation performance stabilized, and the checkpoint with highest validation mIoU was selected for evaluation on the test set. 

\begin{table}[h]
\centering
\caption{Batch sizes per model and dataset.}
\label{tab:batch_sizes}
\begin{tabular}{lrrr}
\toprule
\textbf{Model} & \textbf{\textsc{DALES}} & \textbf{\textsc{FRACTAL}} & \textbf{\textsc{Tracasa-PNA20}} \\
\midrule
KPConv     & 6  & 10 & 10  \\
RandLA-Net & 12 & 10 & 24  \\
PTv3       & 6  & 12 & 10  \\
\bottomrule
\end{tabular}
\end{table}

\begin{table}[t]
\centering
\caption{Class-specific distance thresholds $\tau_c$ (in meters). Values reflect typical object scales and boundary ambiguity zones.}
\label{tab:thresholds}
\begin{tabular}{l l r}
\toprule
\textbf{Dataset} & \textbf{Class} & $\boldsymbol{\tau_c}$ \textbf{(m)} \\
\midrule
\multirow{8}{*}{\textbf{\textsc{DALES}}}
& Ground              & 2.0  \\
& Vegetation          & 3.0  \\
& Buildings           & 10.0 \\
& Cars                & 5.0  \\
& Trucks              & 5.0  \\
& Power Lines         & 5.0  \\
& Fences              & 5.0  \\
& Poles               & 5.0  \\
\midrule
\multirow{7}{*}{\textbf{\textsc{FRACTAL}}}
& Ground              & 2.0  \\
& Vegetation          & 3.0  \\
& Building            & 10.0 \\
& Water               & 10.0 \\
& Bridge              & 5.0  \\
& Permanent Structure & 10.0 \\
& Other               & 10.0 \\
\midrule
\multirow{5}{*}{\textbf{\textsc{Tracasa-PNA20}}}
& Ground              & 2.0  \\
& Low Vegetation      & 2.0  \\
& Med./High Vegetation& 5.0  \\
& Building            & 10.0 \\
& Vehicle             & 5.0  \\
\bottomrule
\end{tabular}
\end{table}

\subsection{Distance Metrics Configuration}

Distance-based metrics require class-specific thresholds $\tau_c$ to clip extreme outliers while preserving meaningful spatial distinctions. Without clipping, isolated errors at extreme distances could mask more persistent error patterns that are critical for model comparison. We established the thresholds in Table~\ref{tab:thresholds} based on the spatial structure and geometric properties of each class, which dictate the range within which errors remain informative.

Classes with continuous spatial structure, such as \textit{ground} and \textit{low vegetation}, form dense layers where misclassifications predominantly occur near boundaries with adjacent classes (e.g., ground--vegetation transitions or building footprints). For these classes, we apply a strict threshold ($\tau_c = 2$\,m), as larger deviations are unlikely to reflect boundary uncertainty and instead correspond to severe spatial confusions.

In contrast, classes representing discrete spatial entities exhibit errors either at object boundaries or as spatially displaced confusions with other structures. For these classes, we set $\tau_c$ according to their typical geometric extent: large structures (\textit{buildings}, \textit{permanent structure}) receive $\tau_c = 10$\,m to account for extended footprints; linear elements (\textit{bridge}, \textit{power lines}) and compact objects (\textit{vehicles}, \textit{poles}, \textit{fences}) use intermediate values ($\tau_c = 5$\,m).

Finally, the \textit{other} class, representing undefined or ambiguous regions, is assigned a conservative maximum threshold ($\tau_c = 10$\,m) to include these points in the evaluation without allowing extreme outliers to skew the overall metrics.

\section{Results} \label{results}

This section presents experimental results across three complementary evaluation perspectives for three state-of-the-art models (KPConv, RandLA-Net, PTv3) evaluated on three aerial LiDAR datasets (\textsc{DALES}, \textsc{FRACTAL} and \textsc{Tracasa-PNA20}). The analysis is organized as follows:

\begin{enumerate}
    \item \textbf{Baseline performance} (Section~\ref{ssec:full_test}): We establish baseline performance using conventional classification metrics (IoU, OA) on full test sets.
    \item \textbf{Hard points evaluation} (Section~\ref{ssec:hard_points}): We measure the impact of restricting evaluation to the hard points subset $\mathcal{H}$ using the same conventional metrics, revealing performance degradation and previously hidden differences in model behavior within challenging regions.
    \item \textbf{Distance-based analysis} (Section~\ref{ssec:distance_analysis}): We apply distance-based metrics (MDE, $\rho$, $\mu$) on $\mathcal{H}$, exposing spatial error patterns invisible to classification-based evaluation.
\end{enumerate}

This analysis demonstrates how our proposed framework provides critical insights into model behavior beyond standard evaluation protocols.

\subsection{Classification-Based Metrics on Full Test Set} \label{ssec:full_test}

We begin by evaluating the three models using standard semantic segmentation metrics on the complete test sets. Tables~\ref{tab:dales_conv}, \ref{tab:fractal_conv}, and \ref{tab:tracasa_conv} present overall accuracy (OA), mean Intersection over Union (mIoU), and per-class IoU for \textsc{DALES}, \textsc{FRACTAL}, and \textsc{Tracasa-PNA20}, respectively. Best values per column are highlighted in bold. These metrics provide a baseline understanding of classification performance before applying our spatially-aware evaluation framework.

Model performance varies substantially across datasets. On \textbf{\textsc{DALES}}, KPConv leads at 83.00\% mIoU, while RandLA-Net (79.66\% mIoU) and PTv3 (79.68\% mIoU) show nearly identical performance. In contrast, \textbf{\textsc{FRACTAL}} reverses this trend: PTv3 achieves the highest mIoU (84.05\%), followed by KPConv (79.78\%) and RandLA-Net (70.56\%). On \textsc{Tracasa-PNA20}, KPConv achieves the highest mIoU (81.02\%), followed closely by RandLA-Net (80.95\% mIoU), while PTv3 ranks third (74.50\% mIoU).

Per-class results reveal consistent challenges across all models and datasets, particularly for minority and geometrically complex classes. In \textbf{\textsc{DALES}}, small objects like \textit{trucks} and thin structures like \textit{fences} achieve only 41.68--48.67\% and 57.85--66.28\% IoU, respectively, compared to $>$92\% for spatially homogeneous classes such as \textit{ground} and \textit{vegetation}. Here, KPConv leads on \textit{fences} (66.28\%) while RandLA-Net and PTv3 achieve comparable but lower scores. On \textbf{\textsc{FRACTAL}}, the \textit{other} class remains problematic at 45.25--58.85\% IoU, while rare classes like \textit{permanent structure} (0.04\% of points) and \textit{bridge} (0.13\% of points) vary widely from 36.59--79.17\% and 51.33--76.77\%, respectively. The gap is most striking for \textit{permanent structure}, where PTv3 reaches 79.17\% compared to 36.59\% for RandLA-Net, reflecting PTv3's stronger handling of rare classes on this dataset. \textbf{\textsc{Tracasa-PNA20}} exhibits similar patterns: dominant classes like \textit{ground} and \textit{building} achieve $>$93\% IoU, while \textit{low vegetation} (1.09\% of points), characterized by ambiguous ground-vegetation boundaries, proves particularly difficult, with IoU ranging from 9.97\% (PTv3) to 33.69\% (KPConv). Cross-model comparison at the per-class level reveals notable differences: KPConv outperforms both competitors on \textit{low vegetation} by a wide margin, suggesting better handling of ambiguous ground-vegetation boundaries.

\input{tables}

\subsection{Classification-Based Metrics on Hard Points Subset} \label{ssec:hard_points}

We now restrict evaluation to the hard points subset $\mathcal{H}$, comprising points misclassified by at least one model (Eq.~\ref{eq:hard_points}). On \textsc{DALES}, $\mathcal{H}$ accounts for 3.80\% of the test set (5.11M out of 134.55M points). \textsc{FRACTAL} exhibits 4.15\% hard points (40.13M out of 966.30M), while \textsc{Tracasa-PNA20} shows 5.51\% (3.10M out of 56.23M).

Tables~\ref{tab:dales_hard_iou}, \ref{tab:fractal_hard_iou}, and \ref{tab:tracasa_hard_iou} report IoU and OA on $\mathcal{H}$. As expected, restricting evaluation to $\mathcal{H}$ leads to substantial drops in absolute metric values, since points correctly classified by all models are excluded by definition: mIoU decreases by about 56 points on \textsc{DALES}, 35 on \textsc{FRACTAL}, and 49 on \textsc{Tracasa-PNA20}, while OA drops from above 95\% on full test sets to 25--70\% on $\mathcal{H}$. These values on $\mathcal{H}$ should not be interpreted as measures of overall model quality. Rather, evaluation on $\mathcal{H}$ focuses attention on the subset where models differ, enabling a finer-grained interpretation of model behavior in challenging regions.

Restricting evaluation to $\mathcal{H}$ naturally amplifies existing differences by removing shared easy points. More importantly, it can also produce per-class IoU ranking inversions. This occurs because $FP_c$ and $FN_c$ are preserved identically in $\mathcal{H}$ (all misclassified points remain by definition), while $TP_c$ is reduced to only those correct predictions at points where at least one other model fails. Since this retained $TP_c$ depends on where the \textit{other} models struggle, a model that dominates on shared easy points may see its per-class advantage reversed when those points are removed:
\begin{itemize}
    \item \textbf{\textsc{DALES}}: On mIoU, RandLA-Net and PTv3 achieve nearly identical scores on the full test set (79.66\% vs 79.68\%), yet RandLA-Net outperforms by 4.6 points on $\mathcal{H}$ (20.24\% vs 15.64\%), a gap representing a 29\% relative difference entirely hidden under the dominance of easy points.
    \item \textbf{\textsc{FRACTAL}}: On \textit{bridge}, PTv3 leads on the full test set (76.77\% vs 71.18\% for KPConv), but KPConv surpasses PTv3 on $\mathcal{H}$ (32.17\% vs 30.71\%), revealing that KPConv handles the difficult \textit{bridge} points more effectively despite lower overall accuracy.
    \item \textbf{\textsc{Tracasa-PNA20}}: On \textit{building}, RandLA-Net achieves the highest IoU on the full test set (96.19\%), narrowly ahead of PTv3 (95.81\%) and KPConv (95.66\%). However, on $\mathcal{H}$, PTv3 rises to first (35.42\%), slightly surpassing RandLA-Net (34.34\%), while KPConv drops sharply to 19.23\%. This ranking shift indicates that the small advantage observed for RandLA-Net on the full test set is not preserved in the most challenging \textit{building} regions, where PTv3 shows greater robustness.
\end{itemize}

Class-specific analysis on $\mathcal{H}$ further reveals asymmetric model-specific weaknesses. When a model's IoU for a given class collapses sharply on $\mathcal{H}$ relative to other models, this indicates that the model contributes disproportionately many errors for that class---errors that the other models classify correctly. For instance, on \textsc{DALES}, PTv3's \textit{vegetation} IoU drops from 92.15\% to 6.95\% on $\mathcal{H}$, while KPConv retains 49.73\%. This asymmetry reveals that a large fraction of PTv3's \textit{vegetation} errors are non-shared: they occur at points where KPConv and RandLA-Net succeed, pointing to model-specific vulnerabilities rather than inherently ambiguous regions. Similar patterns appear on \textsc{FRACTAL}, where RandLA-Net's \textit{building} IoU drops from 86.45\% to 9.19\% on $\mathcal{H}$, while KPConv and PTv3 retain 53.67\% and 61.98\%, respectively, and on \textsc{Tracasa-PNA20}, where PTv3's \textit{med./high vegetation} IoU collapses from 93.73\% to 8.01\%, compared to 42.23\% for KPConv. These patterns are invisible when reporting only mIoU on full test sets, where majority classes achieving $>$93\% IoU dominate the aggregate score.

\subsection{Distance-Based Metrics on Hard Points} \label{ssec:distance_analysis}

We now apply distance-based metrics to characterize spatial error severity. Tables~\ref{tab:dales_dist_hard}, \ref{tab:fractal_dist_hard}, and \ref{tab:tracasa_dist_hard} present per-class MDE (mean distance error in meters), $\rho$ (proportion of distant errors in \%), and $\mu$ (mean distance of non-distant errors in meters).

Model rankings for spatial error severity vary across datasets. On \textbf{\textsc{DALES}}, KPConv achieves the lowest mMDE (1.49~m) compared to RandLA-Net (2.16~m) and PTv3 (2.33~m). However, \textbf{\textsc{FRACTAL}} presents a contrasting pattern: PTv3 achieves the lowest mMDE (1.65~m), substantially outperforming KPConv (2.63~m) and RandLA-Net (3.87~m). Finally, on \textbf{\textsc{Tracasa-PNA20}}, KPConv and RandLA-Net achieve similar mMDE (1.09~m and 1.08~m, respectively), while PTv3 exhibits higher spatial deviations (1.57~m).

Classes representing discrete objects consistently exhibit high proportions of distant errors ($\rho$). On \textbf{\textsc{DALES}}, \textit{trucks} show high proportions, with $\rho$ ranging from 32\% (KPConv) to 59\% (PTv3). On \textbf{\textsc{FRACTAL}}, RandLA-Net struggles particularly on rare classes, where \textit{permanent structure} exhibits $\rho = 99\%$ versus 30\% for PTv3. \textbf{\textsc{Tracasa-PNA20}} shows similar patterns, where PTv3 produces 60\% distant errors for \textit{vehicles}, compared to 25--29\% for KPConv and RandLA-Net. In contrast, continuous classes like \textit{ground} and \textit{vegetation} maintain low boundary deviations ($\mu < 0.6$~m) across all models, indicating that non-distant errors concentrate within narrow transition zones.

Importantly, ranking inversions occur between IoU and distance metrics, revealing that higher classification accuracy does not guarantee better spatial coherence. We highlight two examples across datasets to illustrate this pattern. On \textbf{\textsc{DALES}} \textit{ground}, PTv3 leads in IoU (19.71\% vs 17.92\% for RandLA-Net), but produces higher MDE (0.45~m vs 0.35~m) and $\rho$ (6.63\% vs 3.50\%). Similarly, on \textbf{\textsc{Tracasa-PNA20}} \textit{building}, PTv3 achieves the highest IoU on $\mathcal{H}$ (35.42\% vs 19.23\% for KPConv), yet exhibits higher MDE (2.82~m vs 2.57~m) and $\rho$ (13.51\% vs 8.83\%), indicating more spatially severe errors when misclassifications occur.

\section{Discussion} \label{discussion}

The experimental results presented in Section~\ref{results} demonstrate that the proposed spatially-aware evaluation framework provides critical insights into model behavior that remain invisible under conventional metrics. By combining hard points analysis with distance-based evaluation, we expose three fundamental limitations of standard benchmarking practices: (1) aggregate metrics are dominated by trivially classified points, masking performance differences in challenging regions; (2) classification-based metrics fail to distinguish spatial coherence from classification correctness, hiding geometric error patterns; and (3) model rankings vary substantially across evaluation modes and datasets. This section synthesizes the key findings and discusses their implications for evaluation practices in aerial LiDAR semantic segmentation.

The substantial performance drops observed when restricting evaluation to $\mathcal{H}$, ranging from 35 points to 64 points across models and datasets, confirm that conventional aggregate metrics are overwhelmingly influenced by trivially classified points. On \textsc{DALES}, \textsc{FRACTAL}, and \textsc{Tracasa-PNA20}, only 3.80\%, 4.15\%, and 5.51\% of test points constitute $\mathcal{H}$, respectively, meaning over 94\% of each test set is correctly classified by all models. Yet on \textsc{DALES}, RandLA-Net and PTv3 differ by only 0.02 points in mIoU on the full test (79.66\% vs 79.68\%), but this apparent equivalence disappears on $\mathcal{H}$ where RandLA-Net achieves 20.24\% versus PTv3's 15.64\%, a 4.6 points gap representing 29\% relative difference, exposing substantial robustness differences at class boundaries and geometrically complex regions that remain invisible in aggregate metrics dominated by easy points.

Class-specific analysis on $\mathcal{H}$ further reveals model-specific weaknesses that aggregate metrics obscure. On \textsc{DALES}, PTv3's \textit{vegetation} IoU drops from 92.15\% to 6.95\% on $\mathcal{H}$, while KPConv retains 49.73\%. This pronounced collapse indicates that PTv3 contributes disproportionately to $\mathcal{H}$ for this class: many of its \textit{vegetation} errors occur at points that KPConv and RandLA-Net classify correctly, revealing model-specific vulnerabilities rather than inherently ambiguous regions. Similar patterns appear on \textsc{FRACTAL}, where RandLA-Net's \textit{building} IoU drops from 86.45\% to 9.19\% on $\mathcal{H}$, while KPConv and PTv3 retain 53.67\% and 61.98\%, respectively, and on \textsc{Tracasa-PNA20}, where PTv3's \textit{med./high vegetation} IoU collapses from 93.73\% to 8.01\%, compared to 42.23\% for KPConv. These asymmetric drops across models for the same class provide insight into where each model struggles independently of the others, information that per-class IoU on the full test set cannot reveal due to the dominance of correctly classified points.

Beyond aggregate metric limitations, our distance-based analysis reveals that classification accuracy and spatial error coherence represent distinct and often conflicting dimensions of segmentation quality. The following examples illustrate this pattern across datasets:
\begin{itemize}
    \item \textbf{\textsc{DALES}}: On \textit{ground}, PTv3 achieves the highest IoU on $\mathcal{H}$ (19.71\%) compared to RandLA-Net (17.92\%), suggesting better classification performance in challenging regions. However, distance metrics reverse this ranking: RandLA-Net produces a lower proportion of distant errors ($\rho = 3.50\%$ vs 6.63\%), lower MDE (0.35~m vs 0.45~m), and similar boundary deviations ($\mu = 0.29$~m vs 0.34~m).
    \item \textbf{\textsc{FRACTAL}}: On \textit{bridge}, KPConv achieves slightly higher IoU on $\mathcal{H}$ (32.17\% vs 30.71\% for PTv3), yet exhibits higher MDE (1.82~m vs 1.24~m), a larger proportion of distant errors ($\rho = 20.77\%$ vs 12.62\%), and higher boundary deviations ($\mu = 0.98~m$ vs 0.70~m), demonstrating that better accuracy does not imply better spatial coherence.    
    \item \textbf{\textsc{Tracasa-PNA20}}: On \textit{building}, PTv3 achieves markedly higher IoU on $\mathcal{H}$ (35.42\% vs 19.23\% for KPConv), yet produces higher MDE (2.82~m vs 2.57~m) and a larger proportion of distant errors ($\rho = 13.51\%$ vs 8.83\%), indicating that its misclassifications, although fewer, are more spatially disruptive.
\end{itemize}
The $\mu$ metric complements this information by quantifying boundary-level deviations among non-distant errors. On \textsc{Tracasa-PNA20}, continuous classes such as \textit{ground}, \textit{low vegetation}, and \textit{med./high vegetation} exhibit low boundary deviations, with $\mu$ below 0.6~m for all models, confirming that localized errors concentrate within narrow transition zones. For discrete objects, differences are more pronounced: on \textsc{DALES} \textit{cars}, KPConv achieves $\mu = 0.85$~m versus 1.44~m for PTv3, revealing meaningful spatial extent differences even among non-distant errors.

These observations have practical implications for model selection. In applications requiring spatial consistency, such as DTM generation where misclassified ground points introduce elevation artifacts, models with lower MDE and $\rho$ may be more suitable even at a slight cost in IoU. Conversely, applications focused on overall classification coverage, such as land cover mapping for urban planning, may justify accepting occasional distant errors. Our framework makes these trade-offs explicit, whereas standard evaluation based solely on IoU and OA cannot capture geometric error severity.

\section{Conclusions} \label{conclusions}

This paper introduced a spatially-aware evaluation framework for semantic segmentation of aerial LiDAR point clouds that addresses fundamental limitations of conventional classification-based metrics. We proposed distance-based metrics (MDE, $\rho$, $\mu$) that quantify the geometric severity of misclassifications by measuring how far errors lie from ground-truth class boundaries, and defined the hard points subset $\mathcal{H}$ comprising points misclassified by at least one model to focus evaluation on challenging regions where models exhibit meaningful performance differences and reveal previously hidden model behavior in complex regions.

Experimental validation across three datasets (\textsc{DALES}, \textsc{FRACTAL}, and \textsc{Tracasa-PNA20}) and three architectures (KPConv, RandLA-Net, PTv3) demonstrated that our framework reveals critical distinctions invisible to IoU and overall accuracy. Evaluating on $\mathcal{H}$ not only amplifies differences masked by easy points but can also produce per-class ranking inversions, exposing model-specific strengths and vulnerabilities in challenging regions. Distance-based metrics further reveal that classification accuracy and spatial coherence represent distinct dimensions: models with higher IoU can simultaneously exhibit more spatially severe errors, indicating fundamentally different error behaviors that classification metrics alone cannot capture.

These findings reinforce the need for evaluation frameworks that go beyond classification correctness, particularly in EO applications where spatial consistency directly impacts the quality of derived geospatial products. Our framework makes these trade-offs explicit, enabling application-driven model selection that standard IoU and OA-based evaluation cannot inform.

The proposed framework is straightforward to implement and can be applied to any point cloud segmentation task where spatial error patterns are relevant. The source code implementing all proposed metrics is publicly available\footnote{{\tt\small \url{https://github.com/arin-upna/spatial-eval}}}, facilitating its integration into existing evaluation pipelines.

Several directions remain open for future work. First, extending evaluation to additional architectures and diverse acquisition conditions, such as indoor environments, mobile mapping, or UAV-based surveys, would further validate the generalizability of the framework across different data characteristics. Second, the distance thresholds $\tau_c$ were established based on domain knowledge; investigating adaptive threshold selection based on local point density or application-specific error costs represents a promising refinement. Third, while our framework characterizes point-level spatial errors, extending it to capture object-level spatial coherence could provide additional insights for downstream tasks such as 3D reconstruction or infrastructure mapping. Finally, incorporating distance-based penalties directly into training represents a promising direction, potentially guiding models to produce errors that are not only fewer but also less spatially disruptive.

\section*{Acknowledgment}

Alex Salvatierra holds a predoctoral scholarship funded by the Tracasa Chair in Computer Science and Artificial Intelligence at the Public University of Navarre. This work was also supported by project PID2022-136627NB-I00 (MCIN/\allowbreak AEI/\allowbreak 10.13039/\allowbreak 501100011033/\allowbreak FEDER, EU).

\ifCLASSOPTIONcaptionsoff
  \newpage
\fi

\bibliographystyle{IEEEtran}

\bibliography{main}

\end{document}

%% file: tables.tex
\begin{table*}[t]
    \centering
    \begin{minipage}[t]{0.49\linewidth}
        \centering
               
        \caption{IoU per class, mIoU and OA on \textsc{DALES}.}
        \label{tab:dales_conv}
        \resizebox{\linewidth}{!}{
            \begin{tabular}{l c c c c c c c c c c}
            \toprule
            \multirow{2}{*}{\raisebox{-3.5ex}{Method}} & \multirow{2}{*}{\raisebox{-3.5ex}{OA}} & \multicolumn{9}{c}{IoU} \\ \cmidrule(lr){3-11}
             & & \textit{Mean} & \textit{Ground} & \textit{Vegetation} & \textit{Cars} & \textit{Trucks} & \textit{Power L.} & \textit{Fences} & \textit{Poles} & \textit{Buildings} \\ \midrule
            KPConv & \textbf{98.25} & \textbf{83.00} & \textbf{97.75} & \textbf{95.22} & \textbf{89.13} & \textbf{48.67} & \textbf{96.16} & \textbf{66.28} & \textbf{73.69} & \textbf{97.11} \\
            RandLA-Net & 97.99 & 79.66 & 97.54 & 94.40 & 85.00 & 44.38 & 95.82 & 57.85 & 65.32 & 96.96 \\
            PTv3 & 97.16 & 79.68 & 95.93 & 92.15 & 84.80 & 41.68 & 95.79 & 60.68 & 70.14 & 96.27 \\ 
            \bottomrule
            \end{tabular}
        }

        \vspace{0.625cm}
        
        \caption{IoU per class, mIoU and OA on \textsc{FRACTAL}.}
        \label{tab:fractal_conv}
        \resizebox{\linewidth}{!}{
            \begin{tabular}{l c c c c c c c c c}
            \toprule
            \multirow{2}{*}{\raisebox{-3.5ex}{Method}} & \multirow{2}{*}{\raisebox{-3.5ex}{OA}} & \multicolumn{8}{c}{IoU} \\ \cmidrule(lr){3-10}
             & & \textit{Mean} & \textit{Other} & \textit{Ground} & \textit{Vegetation} & \textit{Building} & \textit{Water} & \textit{Bridge} & \textit{Perm. St.} \\ \midrule
            KPConv & 96.03 & 79.78 & 55.23 & 91.60 & 93.59 & 93.00 & 90.71 & 71.18 & 63.16 \\
            RandLA-Net & 95.56 & 70.56 & 45.25 & 91.06 & 93.18 & 86.45 & 90.07 & 51.33 & 36.59 \\
            PTv3 & \textbf{96.41} & \textbf{84.05} & \textbf{58.85} & \textbf{92.37} & \textbf{94.13} & \textbf{94.12} & \textbf{92.94} & \textbf{76.77} & \textbf{79.17} \\
            \bottomrule
            \end{tabular}
        }

        \vspace{0.2cm}

        \caption{IoU per class, mIoU and OA on \textsc{Tracasa-PNA20}.}
        \label{tab:tracasa_conv}
        \resizebox{\linewidth}{!}{
            \begin{tabular}{l c c c c c c c}
            \toprule
            \multirow{2}{*}{\raisebox{-3.8ex}{Method}} & \multirow{2}{*}{\raisebox{-3.8ex}{OA}} & \multicolumn{6}{c}{IoU} \\ 
            \cmidrule(lr){3-8}
             & & \textit{Mean} & \textit{Ground} & \textit{Low Veg.} & \textit{Med./High Veg.} & \textit{Building} & \textit{Vehicle} \\ \midrule
            KPConv \cite{thomasKPConvFlexibleDeformable2019} & \textbf{97.05} & \textbf{81.02} & \textbf{96.22} & \textbf{33.69} & \textbf{95.28} & 95.66 & 84.25 \\
            RandLA-Net \cite{huRandLANetEfficientSemantic2020} & 96.84 & 80.95 & 96.17 & 31.14 & 94.74 & \textbf{96.19} & \textbf{86.51} \\
            PTv3 \cite{wuPointTransformerV32024} & 96.29 & 74.50 & 95.16 & 9.97 & 93.73 & 95.81 & 77.85 \\ \bottomrule
            \end{tabular}
        }

    \end{minipage}%
    \hfill%
    \begin{minipage}[t]{0.49\linewidth}
        \centering
                
        \caption{IoU per class, mIoU and OA on $\mathcal{H}$ for \textsc{DALES}.}
        \label{tab:dales_hard_iou}
        \resizebox{\linewidth}{!}{
            \begin{tabular}{l c c c c c c c c c c}
            \toprule
            \multirow{2}{*}{\raisebox{-3.5ex}{Method}} & \multirow{2}{*}{\raisebox{-3.5ex}{OA}} & \multicolumn{9}{c}{IoU on $\mathcal{H}$} \\ \cmidrule(lr){3-11}
             & & \textit{Mean} & \textit{Ground} & \textit{Vegetation} & \textit{Cars} & \textit{Trucks} & \textit{Power L.} & \textit{Fences} & \textit{Poles} & \textit{Buildings} \\ \midrule
            KPConv & \textbf{53.91} & \textbf{27.20} & 7.68 & \textbf{49.73} & \textbf{32.41} & 20.54 & \textbf{32.25} & \textbf{18.17} & \textbf{13.03} & \textbf{43.78} \\
            RandLA-Net & 46.98 & 20.24 & 17.92 & 39.56 & 12.80 & \textbf{21.42} & 10.33 & 11.96 & 11.68 & 36.24 \\
            PTv3 & 25.32 & 15.64 & \textbf{19.71} & 6.95 & 16.33 & 17.53 & 20.02 & 12.87 & 10.88 & 20.83 \\ \bottomrule
            \end{tabular}
        }

        \vspace{0.2cm}
        
        \caption{IoU per class, mIoU and OA on $\mathcal{H}$ for \textsc{FRACTAL}.}
        \label{tab:fractal_hard_iou}
        \resizebox{\linewidth}{!}{
            \begin{tabular}{l c c c c c c c c c}
            \toprule
            \multirow{2}{*}{\raisebox{-3.5ex}{Method}} & \multirow{2}{*}{\raisebox{-3.5ex}{OA}} & \multicolumn{8}{c}{IoU on $\mathcal{H}$} \\ \cmidrule(lr){3-10}
             & & \textit{Mean} & \textit{Other} & \textit{Ground} & \textit{Vegetation} & \textit{Building} & \textit{Water} & \textit{Bridge} & \textit{Perm. St.} \\ \midrule
            KPConv & 62.09 & 39.88 & 25.13 & 52.48 & 6.94 & 53.67 & 85.62 & \textbf{32.17} & 23.13 \\
            RandLA-Net & 54.71 & 25.23 & 13.03 & 50.35 & 3.58 & 9.19 & 84.65 & 10.83 & 4.99 \\
            PTv3 & \textbf{70.12} & \textbf{48.87} & \textbf{32.14} & \textbf{62.43} & \textbf{8.41} & \textbf{61.98} & \textbf{89.09} & 30.71 & \textbf{57.31} \\ \bottomrule
            \end{tabular}
        }

        \vspace{0.2cm}

        \caption{IoU per class, mIoU and OA on $\mathcal{H}$ for \textsc{Tracasa-PNA20}.}
        \label{tab:tracasa_hard_iou}
        \resizebox{\linewidth}{!}{
            \begin{tabular}{l c c c c c c c}
            \toprule
            \multirow{2}{*}{\raisebox{-3.8ex}{Method}} & \multirow{2}{*}{\raisebox{-3.8ex}{OA}} & \multicolumn{6}{c}{IoU on $\mathcal{H}$} \\ 
            \cmidrule(lr){3-8}
             & & \textit{Mean} & \textit{Ground} & \textit{Low Veg.} & \textit{Med./High Veg.} & \textit{Building} & \textit{Vehicle} \\ \midrule
            KPConv \cite{thomasKPConvFlexibleDeformable2019} & \textbf{46.48} & \textbf{31.88} & \textbf{28.07} & \textbf{29.72} & \textbf{42.23} & 19.23 & 40.15 \\
            RandLA-Net \cite{huRandLANetEfficientSemantic2020} & 42.61 & 31.29 & 22.43 & 27.73 & 28.38 & 34.34 & \textbf{43.56} \\
            PTv3 \cite{wuPointTransformerV32024} & 32.69 & 16.40 & 27.30 & 4.33 & 8.01 & \textbf{35.42} & 6.94 \\ \bottomrule
            \end{tabular}
        }
        
    \end{minipage}%

    \vspace{0.2cm}
    
    {\setlength{\tabcolsep}{3pt}
    \caption{Distance-based metrics on $\mathcal{H}$ for \textsc{DALES}.}
    \label{tab:dales_dist_hard}
    \resizebox{\textwidth}{!}{
        \begin{tabular}{@{}l c ccc ccc ccc ccc ccc ccc ccc ccc@{}}
        \toprule
         & & \multicolumn{3}{c}{\textbf{Ground}} & \multicolumn{3}{c}{\textbf{Vegetation}} & \multicolumn{3}{c}{\textbf{Cars}} & \multicolumn{3}{c}{\textbf{Trucks}} & \multicolumn{3}{c}{\textbf{Power L.}} & \multicolumn{3}{c}{\textbf{Fences}} & \multicolumn{3}{c}{\textbf{Poles}} & \multicolumn{3}{c}{\textbf{Buildings}} \\
        \cmidrule(lr){3-5} \cmidrule(lr){6-8} \cmidrule(lr){9-11} \cmidrule(lr){12-14} \cmidrule(lr){15-17} \cmidrule(lr){18-20} \cmidrule(lr){21-23} \cmidrule(lr){24-26}
        \textbf{Method} & mMDE & MDE & $\rho$ & $\mu$ & MDE & $\rho$ & $\mu$ & MDE & $\rho$ & $\mu$ & MDE & $\rho$ & $\mu$ & MDE & $\rho$ & $\mu$ & MDE & $\rho$ & $\mu$ & MDE & $\rho$ & $\mu$ & MDE & $\rho$ & $\mu$ \\ \midrule
        KPConv & \textbf{1.49} & 0.46 & 9.29 & 0.30 & \textbf{0.19} & \textbf{1.95} & \textbf{0.13} & \textbf{2.18} & \textbf{31.99} & \textbf{0.85} & \textbf{2.69} & \textbf{32.17} & \textbf{0.18} & \textbf{2.06} & \textbf{10.00} & \textbf{1.18} & \textbf{1.24} & \textbf{29.31} & \textbf{0.50} & \textbf{1.91} & \textbf{30.15} & \textbf{0.58} & \textbf{1.16} & 5.10 & \textbf{0.69} \\
        RandLA-Net & 2.16 & \textbf{0.35} & \textbf{3.50} & \textbf{0.29} & 0.27 & 3.38 & 0.17 & 2.89 & 40.98 & 1.42 & 4.15 & 46.29 & 0.83 & 3.51 & 23.53 & 1.51 & 1.60 & 37.43 & 0.76 & 2.90 & 50.08 & 0.80 & 1.63 & 9.14 & 0.78 \\
        PTv3 & 2.33 & 0.45 & 6.63 & 0.34 & 0.56 & 7.49 & 0.36 & 3.09 & 46.22 & 1.44 & 4.96 & 58.86 & 0.60 & 3.81 & 20.38 & 2.23 & 1.70 & 45.17 & 0.62 & 2.53 & 44.45 & 0.55 & 1.52 & \textbf{4.98} & 1.08 \\ \bottomrule
        \end{tabular}
    }
    }
    
    \vspace{0.2cm}

    {\setlength{\tabcolsep}{3pt}
    \caption{Distance-based metrics on $\mathcal{H}$ for \textsc{FRACTAL}.}
    \label{tab:fractal_dist_hard}
    \resizebox{\textwidth}{!}{
        \begin{tabular}{@{}l c ccc ccc ccc ccc ccc ccc ccc@{}}
        \toprule
         & & \multicolumn{3}{c}{\textbf{Other}} & \multicolumn{3}{c}{\textbf{Ground}} & \multicolumn{3}{c}{\textbf{Vegetation}} & \multicolumn{3}{c}{\textbf{Building}} & \multicolumn{3}{c}{\textbf{Water}} & \multicolumn{3}{c}{\textbf{Bridge}} & \multicolumn{3}{c}{\textbf{Perm. St.}} \\
        \cmidrule(lr){3-5} \cmidrule(lr){6-8} \cmidrule(lr){9-11} \cmidrule(lr){12-14} \cmidrule(lr){15-17} \cmidrule(lr){18-20} \cmidrule(lr){21-23}
        \textbf{Model} & mMDE & MDE & $\rho$ & $\mu$ & MDE & $\rho$ & $\mu$ & MDE & $\rho$ & $\mu$ & MDE & $\rho$ & $\mu$ & MDE & $\rho$ & $\mu$ & MDE & $\rho$ & $\mu$ & MDE & $\rho$ & $\mu$ \\ \midrule
        KPConv & 2.63 & 2.03 & 10.80 & 1.07 & 0.35 & 13.29 & 0.10 & \textbf{1.53} & \textbf{39.35} & 1.23 & 0.93 & 6.32 & \textbf{0.32} & 5.38 & 45.82 & 1.47 & 1.82 & 20.77 & 0.98 & 6.38 & 62.85 & 0.26 \\
        RandLA-Net & 3.87 & 3.62 & 19.24 & 2.10 & 0.52 & 21.60 & 0.12 & 1.56 & 39.60 & 1.27 & 2.42 & 14.54 & 1.13 & 5.28 & \textbf{34.54} & 2.79 & 3.79 & 69.32 & 1.04 & 9.92 & 98.76 & 3.70 \\
        PTv3 & \textbf{1.65} & \textbf{1.63} & \textbf{8.20} & \textbf{0.89} & \textbf{0.28} & \textbf{10.01} & \textbf{0.09} & \textbf{1.53} & 45.43 & \textbf{1.14} & \textbf{0.89} & \textbf{5.98} & \textbf{0.32} & \textbf{2.84} & 21.38 & \textbf{0.90} & \textbf{1.24} & \textbf{12.62} & \textbf{0.70} & \textbf{3.15} & \textbf{30.10} & \textbf{0.20} \\ \bottomrule
        \end{tabular}
    }
    }

    \vspace{0.2cm}

    \caption{Distance-based metrics on $\mathcal{H}$ for \textsc{Tracasa-PNA20}.}
    \label{tab:tracasa_dist_hard}
    \resizebox{\textwidth}{!}{
        \begin{tabular}{@{}l c ccc ccc ccc ccc ccc@{}}
        \toprule
         & & \multicolumn{3}{c}{\textbf{Ground}} & \multicolumn{3}{c}{\textbf{Low Veg.}} & \multicolumn{3}{c}{\textbf{Med./High Veg.}} & \multicolumn{3}{c}{\textbf{Building}} & \multicolumn{3}{c}{\textbf{Vehicle}} \\
        \cmidrule(lr){3-5} \cmidrule(lr){6-8} \cmidrule(lr){9-11} \cmidrule(lr){12-14} \cmidrule(lr){15-17}
        \textbf{Model} & mMDE & MDE & $\rho$ & $\mu$ & MDE & $\rho$ & $\mu$ & MDE & $\rho$ & $\mu$ & MDE & $\rho$ & $\mu$ & MDE & $\rho$ & $\mu$ \\ \midrule
        KPConv & 1.09 & 0.42 & 15.46 & \textbf{0.13} & \textbf{0.25} & \textbf{4.47} & \textbf{0.17} & \textbf{0.44} & \textbf{2.52} & \textbf{0.33} & 2.57 & \textbf{8.83} & 1.85 & 1.74 & 28.83 & 0.42 \\
        RandLA-Net & \textbf{1.08} & \textbf{0.28} & 6.67 & 0.15 & 0.36 & 6.99 & 0.24 & 0.74 & 6.51 & 0.44 & \textbf{2.48} & 13.76 & \textbf{1.28} & \textbf{1.53} & \textbf{25.10} & \textbf{0.37} \\
        PTv3 & 1.57 & \textbf{0.28} & \textbf{3.47} & 0.22 & 0.46 & 4.75 & 0.38 & 0.75 & 5.14 & 0.52 & 2.82 & 13.51 & 1.70 & 3.54 & 60.11 & 1.33 \\ \bottomrule
        \end{tabular}
    }
    
\end{table*}

%% file: main.bib
@inproceedings{armeni3DSemanticParsing2016,
  title = {{{3D Semantic Parsing}} of {{Large-Scale Indoor Spaces}}},
  booktitle = {Proceedings of the {{IEEE Conference}} on {{Computer Vision}} and {{Pattern Recognition}}},
  author = {Armeni, Iro and Sener, Ozan and Zamir, Amir R. and Jiang, Helen and Brilakis, Ioannis and Fischer, Martin and Savarese, Silvio},
  year = 2016,
  pages = {1534--1543},
  urldate = {2025-10-21}
}

@article{behleySemanticKITTIDatasetSemantic2019,
  title = {{{SemanticKITTI}}: {{A Dataset}} for {{Semantic Scene Understanding}} of {{LiDAR Sequences}}},
  author = {Behley, Jens and Garbade, Martin and Milioto, Andres and Quenzel, Jan and Behnke, Sven and Stachniss, Cyrill and Gall, Jurgen},
  year = 2019,
  journal = {2019 IEEECVF Int. Conf. Comput. Vis. ICCV},
  pages = {9296--9306},
  publisher = {IEEE},
  address = {Seoul, Korea (South)},
  doi = {10.1109/ICCV.2019.00939},
  urldate = {2025-10-21},
  copyright = {https://ieeexplore.ieee.org/Xplorehelp/downloads/license-information/IEEE.html}
}

@article{belloReviewDeepLearning2020a,
  title = {Review: {{Deep Learning}} on {{3D Point Clouds}}},
  author = {Bello, Saifullahi Aminu and Yu, Shangshu and Wang, Cheng and Adam, Jibril Muhmmad and Li, Jonathan},
  year = 2020,
  journal = {Remote Sens.},
  volume = {12},
  number = {11},
  pages = {1729},
  publisher = {Multidisciplinary Digital Publishing Institute},
  doi = {10.3390/rs12111729},
  urldate = {2025-10-20},
  copyright = {http://creativecommons.org/licenses/by/3.0/},
  langid = {english}
}

@inproceedings{charlesPointNetDeepLearning2017,
  title = {{{PointNet}}: {{Deep Learning}} on {{Point Sets}} for {{3D Classification}} and {{Segmentation}}},
  booktitle = {2017 {{IEEE Conf}}. {{Comput}}. {{Vis}}. {{Pattern Recognit}}. {{CVPR}}},
  author = {Charles, R. Qi and Su, Hao and Kaichun, Mo and Guibas, Leonidas J.},
  year = 2017,
  pages = {77--85},
  doi = {10.1109/CVPR.2017.16},
  urldate = {2025-09-01}
}

@inproceedings{daiScanNetRichlyAnnotated3D2017,
  title = {{{ScanNet}}: {{Richly-Annotated 3D Reconstructions}} of {{Indoor Scenes}}},
  booktitle = {Proceedings of the {{IEEE Conference}} on {{Computer Vision}} and {{Pattern Recognition}}},
  author = {Dai, Angela and Chang, Angel X. and Savva, Manolis and Halber, Maciej and Funkhouser, Thomas and Niessner, Matthias},
  year = 2017,
  pages = {5828--5839},
  urldate = {2025-10-21}
}

@inproceedings{dekelLargeMarginHierarchical2004,
  title = {Large Margin Hierarchical Classification},
  booktitle = {Proc. {{Twenty-First Int}}. {{Conf}}. {{Mach}}. {{Learn}}.},
  author = {Dekel, Ofer and Keshet, Joseph and Singer, Yoram},
  year = 2004,
  series = {{{ICML}} '04},
  pages = {27},
  publisher = {Association for Computing Machinery},
  address = {New York, NY, USA},
  doi = {10.1145/1015330.1015374},
  urldate = {2025-09-18}
}

@article{druschSentinel2ESAsOptical2012,
  title = {Sentinel-2: {{ESA}}'s {{Optical High-Resolution Mission}} for {{GMES Operational Services}}},
  author = {Drusch, M. and Del Bello, U. and Carlier, S. and Colin, O. and Fernandez, V. and Gascon, F. and Hoersch, B. and Isola, C. and Laberinti, P. and Martimort, P. and Meygret, A. and Spoto, F. and Sy, O. and Marchese, F. and Bargellini, P.},
  year = 2012,
  journal = {Remote Sensing of Environment},
  series = {The {{Sentinel Missions}} - {{New Opportunities}} for {{Science}}},
  volume = {120},
  pages = {25--36},
  doi = {10.1016/j.rse.2011.11.026},
  urldate = {2026-01-23}
}

@inproceedings{elkanFoundationsCostsensitiveLearning2001,
  title = {The Foundations of Cost-Sensitive Learning},
  booktitle = {Proc. 17th {{Int}}. {{Jt}}. {{Conf}}. {{Artif}}. {{Intell}}. - {{Vol}}. 2},
  author = {Elkan, Charles},
  year = 2001,
  series = {{{IJCAI}}'01},
  pages = {973--978},
  publisher = {Morgan Kaufmann Publishers Inc.},
  address = {San Francisco, CA, USA},
  doi = {10.48550/arXiv.1911.08287},
  urldate = {2025-09-01}
}

@misc{gaydonFRACTALUltraLargeScaleAerial2024,
  title = {{{FRACTAL}}: {{An Ultra-Large-Scale Aerial Lidar Dataset}} for {{3D Semantic Segmentation}} of {{Diverse Landscapes}}},
  author = {Gaydon, Charles and Daab, Michel and Roche, Floryne},
  year = 2024,
  number = {arXiv:2405.04634},
  eprint = {2405.04634},
  primaryclass = {cs},
  publisher = {arXiv},
  doi = {10.48550/arXiv.2405.04634},
  urldate = {2025-09-01},
  archiveprefix = {arXiv}
}

@misc{GoogledeepmindSurfacedistance2018,
  title = {Google-Deepmind/Surface-Distance},
  year = 2018,
  urldate = {2025-09-02},
  copyright = {Apache-2.0},
  howpublished = {Google DeepMind}
}

@article{guoDeepLearning3D2021a,
  title = {Deep {{Learning}} for {{3D Point Clouds}}: {{A Survey}}},
  author = {Guo, Yulan and Wang, Hanyun and Hu, Qingyong and Liu, Hao and Liu, Li and Bennamoun, Mohammed},
  year = 2021,
  journal = {IEEE Trans. Pattern Anal. Mach. Intell.},
  volume = {43},
  number = {12},
  pages = {4338--4364},
  doi = {10.1109/TPAMI.2020.3005434},
  urldate = {2025-10-20}
}

@article{hackelSEMANTIC3DNETNEWLARGESCALE2017,
  title = {{{SEMANTIC3D}}.{{NET}}: {{A NEW LARGE-SCALE POINT CLOUD CLASSIFICATION BENCHMARK}}},
  author = {Hackel, T. and Savinov, N. and Ladicky, L. and Wegner, J. D. and Schindler, K. and Pollefeys, M.},
  year = 2017,
  journal = {ISPRS Ann. Photogramm. Remote Sens. Spat. Inf. Sci.},
  volume = {IV-1-W1},
  pages = {91--98},
  publisher = {Copernicus GmbH},
  doi = {10.5194/isprs-annals-IV-1-W1-91-2017},
  urldate = {2025-10-21},
  langid = {english}
}

@inproceedings{huRandLANetEfficientSemantic2020,
  title = {{{RandLA-Net}}: {{Efficient Semantic Segmentation}} of {{Large-Scale Point Clouds}}},
  booktitle = {2020 {{IEEECVF Conf}}. {{Comput}}. {{Vis}}. {{Pattern Recognit}}. {{CVPR}}},
  author = {Hu, Qingyong and Yang, Bo and Xie, Linhai and Rosa, Stefano and Guo, Yulan and Wang, Zhihua and Trigoni, Niki and Markham, Andrew},
  year = 2020,
  pages = {11105--11114},
  doi = {10.1109/CVPR42600.2020.01112},
  urldate = {2025-09-01}
}

@article{huttenlocherComparingImagesUsing1993,
  title = {Comparing Images Using the {{Hausdorff}} Distance},
  author = {Huttenlocher, D.P. and Klanderman, G.A. and Rucklidge, W.J.},
  year = 1993,
  journal = {IEEE Trans. Pattern Anal. Mach. Intell.},
  volume = {15},
  number = {9},
  pages = {850--863},
  doi = {10.1109/34.232073},
  urldate = {2025-09-02}
}

@article{kharroubiThreeDimensionalChange2022,
  title = {Three {{Dimensional Change Detection Using Point Clouds}}: {{A Review}}},
  author = {Kharroubi, Abderrazzaq and Poux, Florent and Ballouch, Zouhair and Hajji, Rafika and Billen, Roland},
  year = 2022,
  journal = {Geomatics},
  volume = {2},
  number = {4},
  pages = {457--485},
  publisher = {Multidisciplinary Digital Publishing Institute},
  doi = {10.3390/geomatics2040025},
  urldate = {2026-01-23},
  copyright = {http://creativecommons.org/licenses/by/3.0/},
  langid = {english}
}

@article{knapitschTanksTemplesBenchmarking2017,
  title = {Tanks and Temples: Benchmarking Large-Scale Scene Reconstruction},
  author = {Knapitsch, Arno and Park, Jaesik and Zhou, Qian-Yi and Koltun, Vladlen},
  year = 2017,
  journal = {ACM Trans. Graph.},
  volume = {36},
  number = {4},
  pages = {78:1--78:13},
  doi = {10.1145/3072959.3073599},
  urldate = {2025-09-02}
}

@article{lecunDeepLearning2015,
  title = {Deep Learning},
  author = {LeCun, Yann and Bengio, Yoshua and Hinton, Geoffrey},
  year = 2015,
  journal = {Nature},
  volume = {521},
  number = {7553},
  pages = {436--444},
  publisher = {Nature Publishing Group},
  doi = {10.1038/nature14539},
  urldate = {2026-01-23},
  copyright = {2015 Springer Nature Limited},
  langid = {english}
}

@article{liDeepLearningLiDAR2021,
  title = {Deep {{Learning}} for {{LiDAR Point Clouds}} in {{Autonomous Driving}}: {{A Review}}},
  author = {Li, Ying and Ma, Lingfei and Zhong, Zilong and Liu, Fei and Chapman, Michael A. and Cao, Dongpu and Li, Jonathan},
  year = 2021,
  journal = {IEEE Trans. Neural Netw. Learn. Syst.},
  volume = {32},
  number = {8},
  pages = {3412--3432},
  doi = {10.1109/TNNLS.2020.3015992},
  urldate = {2025-09-01}
}

@misc{linFocalLossDense2018,
  title = {Focal {{Loss}} for {{Dense Object Detection}}},
  author = {Lin, Tsung-Yi and Goyal, Priya and Girshick, Ross and He, Kaiming and Doll{\'a}r, Piotr},
  year = 2018,
  number = {arXiv:1708.02002},
  eprint = {1708.02002},
  primaryclass = {cs},
  publisher = {arXiv},
  doi = {10.48550/arXiv.1708.02002},
  urldate = {2025-09-18},
  archiveprefix = {arXiv}
}

@inproceedings{liPointCNNConvolutionXTransformed2018,
  title = {{{PointCNN}}: {{Convolution On X-Transformed Points}}},
  booktitle = {Adv. {{Neural Inf}}. {{Process}}. {{Syst}}.},
  author = {Li, Yangyan and Bu, Rui and Sun, Mingchao and Wu, Wei and Di, Xinhan and Chen, Baoquan},
  year = 2018,
  volume = {31},
  publisher = {Curran Associates, Inc.},
  doi = {10.48550/arXiv.1801.07791},
  urldate = {2025-09-01}
}

@book{liu3DPointCloud2021,
  title = {{{3D Point Cloud Analysis}}: {{Traditional}}, {{Deep Learning}}, and {{Explainable Machine Learning Methods}}},
  author = {Liu, Shan and Zhang, Min and Kadam, Pranav and Kuo, C.-C. Jay},
  year = 2021,
  publisher = {Springer International Publishing},
  address = {Cham},
  doi = {10.1007/978-3-030-89180-0},
  urldate = {2026-01-23},
  copyright = {https://www.springer.com/tdm},
  langid = {english}
}

@article{liuAirborneLiDARGeneration2008,
  title = {Airborne {{LiDAR}} for {{DEM}} Generation: Some Critical Issues},
  author = {Liu, Xiaoye},
  year = 2008,
  journal = {Prog. Phys. Geogr. Earth Environ.},
  volume = {32},
  number = {1},
  pages = {31--49},
  publisher = {SAGE Publications Ltd},
  doi = {10.1177/0309133308089496},
  urldate = {2026-01-23},
  langid = {english}
}

@inproceedings{luSimplifiedMarkovRandom2012,
  title = {Simplified Markov Random Fields for Efficient Semantic Labeling of {{3D}} Point Clouds},
  booktitle = {2012 {{IEEERSJ Int}}. {{Conf}}. {{Intell}}. {{Robots Syst}}.},
  author = {Lu, Yan and Rasmussen, Christopher},
  year = 2012,
  pages = {2690--2697},
  doi = {10.1109/IROS.2012.6386039},
  urldate = {2026-01-23}
}

@article{niemeyerCONDITIONALRANDOMFIELDS2012,
  title = {{{CONDITIONAL RANDOM FIELDS FOR LIDAR POINT CLOUD CLASSIFICATION IN COMPLEX URBAN AREAS}}},
  author = {Niemeyer, J. and Rottensteiner, F. and Soergel, U.},
  year = 2012,
  journal = {ISPRS Ann. Photogramm. Remote Sens. Spat. Inf. Sci.},
  volume = {I-3},
  pages = {263--268},
  publisher = {Copernicus GmbH},
  doi = {10.5194/isprsannals-I-3-263-2012},
  urldate = {2026-01-23},
  langid = {english}
}

@inproceedings{oksuzLocalizationRecallPrecision2018,
  title = {Localization {{Recall Precision}} ({{LRP}}): {{A New Performance Metric}} for {{Object Detection}}},
  booktitle = {Comput. {{Vis}}. -- {{ECCV}} 2018 15th {{Eur}}. {{Conf}}. {{Munich Ger}}. {{Sept}}. 8--14 2018 {{Proc}}. {{Part VII}}},
  author = {Oksuz, Kemal and Cam, Baris Can and Akbas, Emre and Kalkan, Sinan},
  year = 2018,
  pages = {521--537},
  publisher = {Springer-Verlag},
  address = {Berlin, Heidelberg},
  doi = {10.1007/978-3-030-01234-2_31},
  urldate = {2025-09-01}
}

@article{paredesLearningWeightedMetrics2006,
  title = {Learning Weighted Metrics to Minimize Nearest-Neighbor Classification Error},
  author = {Paredes, Roberto and Vidal, Enrique},
  year = 2006,
  journal = {IEEE Trans Pattern Anal Mach Intell},
  volume = {28},
  number = {7},
  pages = {1100--1110},
  doi = {10.1109/TPAMI.2006.145},
  langid = {english}
}

@inproceedings{qiPointNetDeepHierarchical2017,
  title = {{{PointNet}}++: {{Deep Hierarchical Feature Learning}} on {{Point Sets}} in a {{Metric Space}}},
  booktitle = {Adv. {{Neural Inf}}. {{Process}}. {{Syst}}.},
  author = {Qi, Charles Ruizhongtai and Yi, Li and Su, Hao and Guibas, Leonidas J},
  year = 2017,
  volume = {30},
  publisher = {Curran Associates, Inc.},
  doi = {10.48550/arXiv.1706.02413},
  urldate = {2025-09-01}
}

@article{reutebuchLightDetectionRanging2005,
  title = {Light Detection and Ranging ({{LIDAR}}): An Emerging Tool for Multiple Resource Inventory.},
  author = {Reutebuch, Stephen E. and Andersen, Hans-Erik and McGaughey, Robert J.},
  year = 2005,
  journal = {J. For. 286-292},
  urldate = {2026-01-23},
  langid = {english}
}

@article{singhDeepLearningbasedSemantic2024,
  title = {Deep Learning-Based Semantic Segmentation of Three-Dimensional Point Cloud: A Comprehensive Review},
  author = {Singh, Dheerendra Pratap and Yadav, Manohar},
  year = 2024,
  journal = {Int. J. Remote Sens.},
  volume = {45},
  number = {2},
  pages = {532--586},
  publisher = {Taylor \& Francis},
  doi = {10.1080/01431161.2023.2297177},
  urldate = {2025-09-01}
}

@inproceedings{sunHierarchicalTextClassification2001,
  title = {Hierarchical Text Classification and Evaluation},
  booktitle = {Proc. 2001 {{IEEE Int}}. {{Conf}}. {{Data Min}}.},
  author = {Sun, Aixin and Lim, Ee-Peng},
  year = 2001,
  pages = {521--528},
  doi = {10.1109/ICDM.2001.989560},
  urldate = {2025-09-18}
}

@inproceedings{thomasKPConvFlexibleDeformable2019,
  title = {{{KPConv}}: {{Flexible}} and {{Deformable Convolution}} for {{Point Clouds}}},
  booktitle = {2019 {{IEEECVF Int}}. {{Conf}}. {{Comput}}. {{Vis}}. {{ICCV}}},
  author = {Thomas, Hugues and Qi, Charles R. and Deschaud, Jean-Emmanuel and Marcotegui, Beatriz and Goulette, Fran{\c c}ois and Guibas, Leonidas},
  year = 2019,
  pages = {6410--6419},
  doi = {10.1109/ICCV.2019.00651},
  urldate = {2025-09-01}
}

@misc{varneyDALESLargescaleAerial2020,
  title = {{{DALES}}: {{A Large-scale Aerial LiDAR Data Set}} for {{Semantic Segmentation}}},
  author = {Varney, Nina and Asari, Vijayan K. and Graehling, Quinn},
  year = 2020,
  number = {arXiv:2004.11985},
  eprint = {2004.11985},
  primaryclass = {cs},
  publisher = {arXiv},
  doi = {10.48550/arXiv.2004.11985},
  urldate = {2025-09-01},
  archiveprefix = {arXiv}
}

@article{wangLiDARPointClouds2018,
  title = {{{LiDAR Point Clouds}} to 3-{{D Urban Models}}: {{A Review}}},
  author = {Wang, Ruisheng and Peethambaran, Jiju and Chen, Dong},
  year = 2018,
  journal = {IEEE J. Sel. Top. Appl. Earth Obs. Remote Sens.},
  volume = {11},
  number = {2},
  pages = {606--627},
  doi = {10.1109/JSTARS.2017.2781132},
  urldate = {2025-11-26}
}

@article{winkerOverviewCALIPSOMission2009,
  title = {Overview of the {{CALIPSO Mission}} and {{CALIOP Data Processing Algorithms}}},
  author = {Winker, David M. and Vaughan, Mark A. and Omar, Ali and Hu, Yongxiang and Powell, Kathleen A. and Liu, Zhaoyan and Hunt, William H. and Young, Stuart A.},
  year = 2009,
  journal = {J. Atmospheric Ocean. Technol.},
  volume = {26},
  number = {11},
  pages = {2310--2323},
  publisher = {American Meteorological Society},
  doi = {10.1175/2009JTECHA1281.1},
  urldate = {2026-01-23},
  chapter = {Journal of Atmospheric and Oceanic Technology},
  langid = {english}
}

@article{wulderLidarSamplingLargearea2012,
  title = {Lidar Sampling for Large-Area Forest Characterization: {{A}} Review},
  author = {Wulder, Michael A. and White, Joanne C. and Nelson, Ross F. and N{\ae}sset, Erik and {\O}rka, Hans Ole and Coops, Nicholas C. and Hilker, Thomas and Bater, Christopher W. and Gobakken, Terje},
  year = 2012,
  journal = {Remote Sensing of Environment},
  volume = {121},
  pages = {196--209},
  doi = {10.1016/j.rse.2012.02.001},
  urldate = {2025-11-26}
}

@inproceedings{wuPointTransformerV32024,
  title = {Point {{Transformer V3}}: {{Simpler}}, {{Faster}}, {{Stronger}}},
  booktitle = {2024 {{IEEECVF Conf}}. {{Comput}}. {{Vis}}. {{Pattern Recognit}}. {{CVPR}}},
  author = {Wu, Xiaoyang and Jiang, Li and Wang, Peng-Shuai and Liu, Zhijian and Liu, Xihui and Qiao, Yu and Ouyang, Wanli and He, Tong and Zhao, Hengshuang},
  year = 2024,
  pages = {4840--4851},
  doi = {10.1109/CVPR52733.2024.00463},
  urldate = {2025-09-01}
}

@article{yanUrbanLandCover2015,
  title = {Urban Land Cover Classification Using Airborne {{LiDAR}} Data: {{A}} Review},
  author = {Yan, Wai Yeung and Shaker, Ahmed and {El-Ashmawy}, Nagwa},
  year = 2015,
  journal = {Remote Sensing of Environment},
  volume = {158},
  pages = {295--310},
  doi = {10.1016/j.rse.2014.11.001},
  urldate = {2025-11-26}
}

@article{yooYUTOSEMANTICLARGE2023,
  title = {{{YUTO SEMANTIC}}: {{A LARGE SCALE AERIAL LIDAR DATASET FOR SEMANTIC SEGMENTATION}}},
  author = {Yoo, S. and Ko, C. and Sohn, G. and Lee, H.},
  year = 2023,
  journal = {Int. Arch. Photogramm. Remote Sens. Spat. Inf. Sci.},
  volume = {XLVIII-1-W2-2023},
  pages = {209--215},
  publisher = {Copernicus GmbH},
  doi = {10.5194/isprs-archives-XLVIII-1-W2-2023-209-2023},
  urldate = {2025-09-01},
  langid = {english}
}

@article{yuAutomatedDerivationUrban2010,
  title = {Automated Derivation of Urban Building Density Information Using Airborne {{LiDAR}} Data and Object-Based Method},
  author = {Yu, Bailang and Liu, Hongxing and Wu, Jianping and Hu, Yingjie and Zhang, Li},
  year = 2010,
  journal = {Landscape and Urban Planning},
  series = {Climate {{Change}} and {{Spatial Planning}}},
  volume = {98},
  number = {3},
  pages = {210--219},
  doi = {10.1016/j.landurbplan.2010.08.004},
  urldate = {2026-01-23}
}

@article{zhangDeepLearningbased3D2023,
  title = {Deep Learning-Based {{3D}} Point Cloud Classification: {{A}} Systematic Survey and Outlook},
  author = {Zhang, Huang and Wang, Changshuo and Tian, Shengwei and Lu, Baoli and Zhang, Liping and Ning, Xin and Bai, Xiao},
  year = 2023,
  journal = {Displays},
  volume = {79},
  pages = {102456},
  doi = {10.1016/j.displa.2023.102456},
  urldate = {2025-09-01}
}

@article{zhangEasytoUseAirborneLiDAR2016,
  title = {An {{Easy-to-Use Airborne LiDAR Data Filtering Method Based}} on {{Cloth Simulation}}},
  author = {Zhang, Wuming and Qi, Jianbo and Wan, Peng and Wang, Hongtao and Xie, Donghui and Wang, Xiaoyan and Yan, Guangjian},
  year = 2016,
  journal = {Remote Sens.},
  volume = {8},
  number = {6},
  pages = {501},
  publisher = {Multidisciplinary Digital Publishing Institute},
  doi = {10.3390/rs8060501},
  urldate = {2026-01-23},
  copyright = {http://creativecommons.org/licenses/by/3.0/},
  langid = {english}
}

@article{zhangReviewDeepLearningBased2019,
  title = {A {{Review}} of {{Deep Learning-Based Semantic Segmentation}} for {{Point Cloud}}},
  author = {Zhang, Jiaying and Zhao, Xiaoli and Chen, Zheng and Lu, Zhejun},
  year = 2019,
  journal = {IEEE Access},
  volume = {7},
  pages = {179118--179133},
  doi = {10.1109/ACCESS.2019.2958671},
  urldate = {2026-01-23}
}

@article{zhangSVMBasedClassificationSegmented2013,
  title = {{{SVM-Based Classification}} of {{Segmented Airborne LiDAR Point Clouds}} in {{Urban Areas}}},
  author = {Zhang, Jixian and Lin, Xiangguo and Ning, Xiaogang},
  year = 2013,
  journal = {Remote Sens.},
  volume = {5},
  number = {8},
  pages = {3749--3775},
  publisher = {Multidisciplinary Digital Publishing Institute},
  doi = {10.3390/rs5083749},
  urldate = {2026-01-23},
  copyright = {http://creativecommons.org/licenses/by/3.0/},
  langid = {english}
}

@article{zhengDistanceIoULossFaster2020,
  title = {Distance-{{IoU Loss}}: {{Faster}} and {{Better Learning}} for {{Bounding Box Regression}}},
  author = {Zheng, Zhaohui and Wang, Ping and Liu, Wei and Li, Jinze and Ye, Rongguang and Ren, Dongwei},
  year = 2020,
  journal = {AAAI},
  volume = {34},
  number = {07},
  pages = {12993--13000},
  doi = {10.1609/aaai.v34i07.6999},
  urldate = {2025-09-02},
  copyright = {https://www.aaai.org},
  langid = {english}
}

@inproceedings{robertEfficient3DSemantic2023,
  title = {Efficient {{3D Semantic Segmentation}} with {{Superpoint Transformer}}},
  booktitle = {2023 {{IEEE}}/{{CVF International Conference}} on {{Computer Vision}} ({{ICCV}})},
  author = {Robert, Damien and Raguet, Hugo and Landrieu, Loic},
  year = 2023,
  pages = {17149--17158},
  doi = {10.1109/ICCV51070.2023.01577},
  urldate = {2025-09-01}
}

@article{wuPointTransformerV22022,
  title = {Point {{Transformer V2}}: {{Grouped Vector Attention}} and {{Partition-based Pooling}}},
  author = {Wu, Xiaoyang and Lao, Yixing and Jiang, Li and Liu, Xihui and Zhao, Hengshuang},
  year = 2022,
  journal = {Advances in Neural Information Processing Systems},
  volume = {35},
  pages = {33330--33342},
  doi = {10.48550/arXiv.2210.05666},
  urldate = {2025-09-01},
  langid = {english}
}

@inproceedings{zhaoPointTransformer2021,
  title = {Point {{Transformer}}},
  booktitle = {2021 {{IEEE}}/{{CVF International Conference}} on {{Computer Vision}} ({{ICCV}})},
  author = {Zhao, Hengshuang and Jiang, Li and Jia, Jiaya and Torr, Philip and Koltun, Vladlen},
  year = 2021,
  pages = {16239--16248},
  doi = {10.1109/ICCV48922.2021.01595},
  urldate = {2025-09-01}
}

@misc{melekhovECLAIRHighFidelityAerial2024,
  title = {{{ECLAIR}}: {{A High-Fidelity Aerial LiDAR Dataset}} for {{Semantic Segmentation}}},
  author = {Melekhov, Iaroslav and Umashankar, Anand and Kim, Hyeong-Jin and Serkov, Vladislav and Argyle, Dusty},
  year = 2024,
  number = {arXiv:2404.10699},
  eprint = {2404.10699},
  primaryclass = {cs},
  publisher = {arXiv},
  doi = {10.48550/arXiv.2404.10699},
  urldate = {2025-09-01},
  archiveprefix = {arXiv}
}
